%% file: main.tex
\documentclass[sigconf]{acmart}
\usepackage{makecell}
\usepackage{algpseudocode}
\usepackage{threeparttable}
\usepackage{booktabs}
\usepackage{graphicx}
\usepackage{subcaption}
\usepackage{tablefootnote} 
\usepackage {colortbl}
\usepackage{multirow}
\usepackage{array}
\newcolumntype{P}[1]{>{\centering\arraybackslash}p{#1}}
\usepackage{longtable}

\usepackage{algpseudocode} 
\usepackage{algorithm}
\definecolor{lightred}{rgb}{1,0.8,0.8}
\definecolor{lightgreen}{rgb}{0.8,1,0.8}

\AtBeginDocument{
  }

\setcopyright{acmlicensed}
\copyrightyear{2025}
\acmYear{2025}
\acmDOI{XXXXXXX.XXXXXXX}

\acmISBN{978-1-4503-XXXX-X/2018/06}

\title{{\raisebox{-0.14\height}{\includegraphics[height=1.1em]{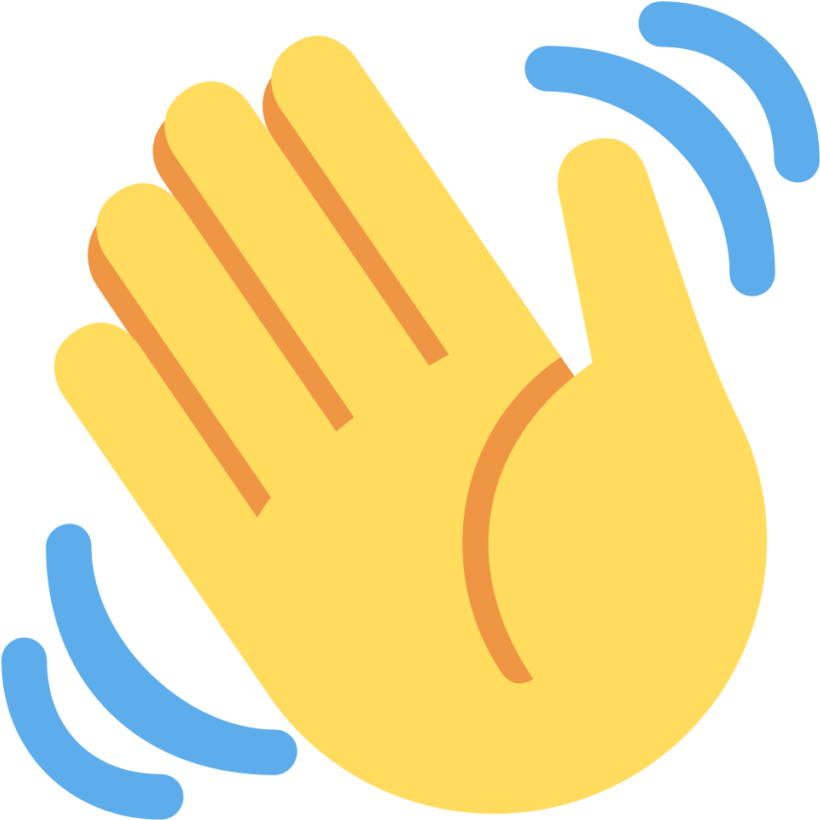}}\hspace{0.2em} HiBench: Benchmarking LLMs Capability on Hierarchical Structure Reasoning}
}

\author{Zhuohang Jiang}
\authornote{Authors contributed equally to this research.}
\email{zhuohang.jiang@connect.polyu.hk}
\affiliation{%
  \institution{The Hong Kong Polytechnic University}
  \country{Hong Kong}
}

\author{Pangjing Wu}
\authornotemark[1]
\email{pang-jing.wu@connect.polyu.hk}
\affiliation{%
  \institution{The Hong Kong Polytechnic University}
  \country{Hong Kong}
}

\author{Ziran Liang}
\authornotemark[1]
\email{ziran.liang@connect.polyu.hk}
\affiliation{%
  \institution{The Hong Kong Polytechnic University}
  \country{Hong Kong}
}

\author{Peter Q. Chen}
\authornotemark[1]
\email{peter-q.chen@connect.polyu.hk}
\affiliation{%
  \institution{The Hong Kong Polytechnic University}
  \country{Hong Kong}
}

\author{Xu Yuan}
\authornotemark[1]
\email{xander.yuan@connect.polyu.hk}
\affiliation{%
  \institution{The Hong Kong Polytechnic University}
  \country{Hong Kong}
}

\author{Ye Jia}
\authornotemark[1]
\email{ye-aimmeng.jia@connect.polyu.hk}
\affiliation{%
  \institution{The Hong Kong Polytechnic University}
  \country{Hong Kong}
}

\author{Jiancheng Tu}
\authornotemark[1]
\email{jiancheng.tu@connect.polyu.hk}
\affiliation{%
  \institution{The Hong Kong Polytechnic University}
  \country{Hong Kong}
}

\author{Chen Li}
\email{richard-chen.li@polyu.edu.hk}
\affiliation{%
  \institution{The Hong Kong Polytechnic University}
  \country{Hong Kong}
}

\author{Peter H.F. Ng}
\email{peter.nhf@polyu.edu.hk}
\affiliation{%
  \institution{The Hong Kong Polytechnic University}
  \country{Hong Kong}
}

\author{Qing Li}
\authornote{Corresponding Author.}
\email{csqli@comp.polyu.edu.hk}
\affiliation{%
  \institution{The Hong Kong Polytechnic University}
  \country{Hong Kong}
}

\renewcommand{\shortauthors}{Jiang et al.}

\begin{document}

\begin{abstract}
Structure reasoning is a fundamental capability of large language models (LLMs), enabling them to reason about structured commonsense and answer multi-hop questions. 
However, existing benchmarks for structure reasoning mainly focus on horizontal and coordinate structures (\emph{e.g.} graphs), overlooking the hierarchical relationships within them. 
Hierarchical structure reasoning is crucial for human cognition, particularly in memory organization and problem-solving. It also plays a key role in various real-world tasks, such as information extraction and decision-making. 
To address this gap, we propose HiBench, the first framework spanning from initial structure generation to final proficiency assessment, designed to benchmark the hierarchical reasoning capabilities of LLMs systematically.
HiBench encompasses six representative scenarios, covering both fundamental and practical aspects, and consists of 30 tasks with varying hierarchical complexity, totaling 39,519 queries.
To evaluate LLMs comprehensively, we develop five capability dimensions that depict different facets of hierarchical structure understanding. 
Through extensive evaluation of 20 LLMs from 10 model families, we reveal key insights into their capabilities and limitations: 1) existing LLMs show proficiency in basic hierarchical reasoning tasks; 2) they still struggle with more complex structures and implicit hierarchical representations, especially in structural modification and textual reasoning. 
Based on these findings, we create a small yet well-designed instruction dataset, which enhances LLMs' performance on HiBench by an average of 88.84\% (Llama-3.1-8B) and 31.38\% (Qwen2.5-7B) across all tasks.
The HiBench dataset and toolkit are available  \href{https://github.com/jzzzzh/HiBench}{here} to encourage evaluation.
\end{abstract}



\maketitle

\input{1-intro}
\input{2-related-work}
\input{3-hibench}
\input{5-results}
\input{6-conclusion}

\bibliographystyle{ACM-Reference-Format}

\newpage
\appendix
\input{9-appendix}

\end{document}

%% file: 1-intro.tex
\section{Introduction}
Recently, Large Language Models (LLMs) have shown remarkable performance across a variety of tasks, such as conversational AI~\cite{meshram2021conversational,bettayeb2024exploring, casheekar2024contemporary}, text summarization~\cite{jin2024comprehensive, zhang2024systematic}, language translation~\cite{koshkin2024transllama, lu2024llamax, gong2024llms}, programming assistance~\cite{liu2024exploring, nam2024using}.
These advancements have driven substantial progress in practical applications, including healthcare~\cite{qiu2024llm, mirzaei2024clinician, goyal2024healai}, finance~\cite{yu2025fincon, xing2025designing, zhao2024revolutionizing}, and software development~\cite{manish2024autonomous, lin2024llm, xia2024agentless}.
Notably, the emergent cognitive abilities in LLMs have been observed to increasingly mirror certain aspects of human intelligence, which suggests potential parallels between LLMs' behavior and human cognitive processes, arousing discussions about possible pathways to artificial general intelligence (AGI)~\cite{zhong2024evaluation, feng2024far}.
One fundamental principle of human cognition is hierarchical reasoning, which is essential for memory organization, problem-solving, and decision-making~\cite{cogscience_3,cogscience_2,cogscience_1}, allowing humans to understand and organize complex relationships with structured knowledge effectively.
Consequently, assessing whether and how well hierarchical reasoning is exhibited in LLMs is significant for further investigating the alignment between LLMs' capabilities and human cognition.
 
While numerous benchmarks have been developed to assess various cognitive capabilities of LLMs, such as memory retrieval\cite{kagaya2024rap}, commonsense understanding\cite{kwon2024toward, perak2024incorporating}, and structure reasoning\cite{zhou2025self, musker2024semantic}, there remains a significant gap when it comes to evaluating hierarchical reasoning.
Existing structure reasoning benchmarks for LLMs focus primarily on tasks involving horizontal or coordinate structures, such as graph~\cite{dai2024revisiting, dai2024large, wang2023can} or table~\cite{sui2024table, wang2024chain}, but overlook the critical hierarchical nature of cognitive reasoning, where information is processed at different levels of abstraction. This absence hinders a comprehensive understanding of LLMs' true cognitive abilities. 
To fill the essential gap in current evaluation frameworks for LLMs, we propose \textbf{HiBench}, the first benchmark specifically designed to evaluate LLMs' capability on hierarchical structure reasoning, providing a systematic framework for assessing how LLMs organize, process, and reason with multi-level information across multiple capability dimensions and scenarios.

\begin{figure}[t]
\centering
\begin{subfigure}[a]{1\linewidth}
    \includegraphics[width=0.95\linewidth]{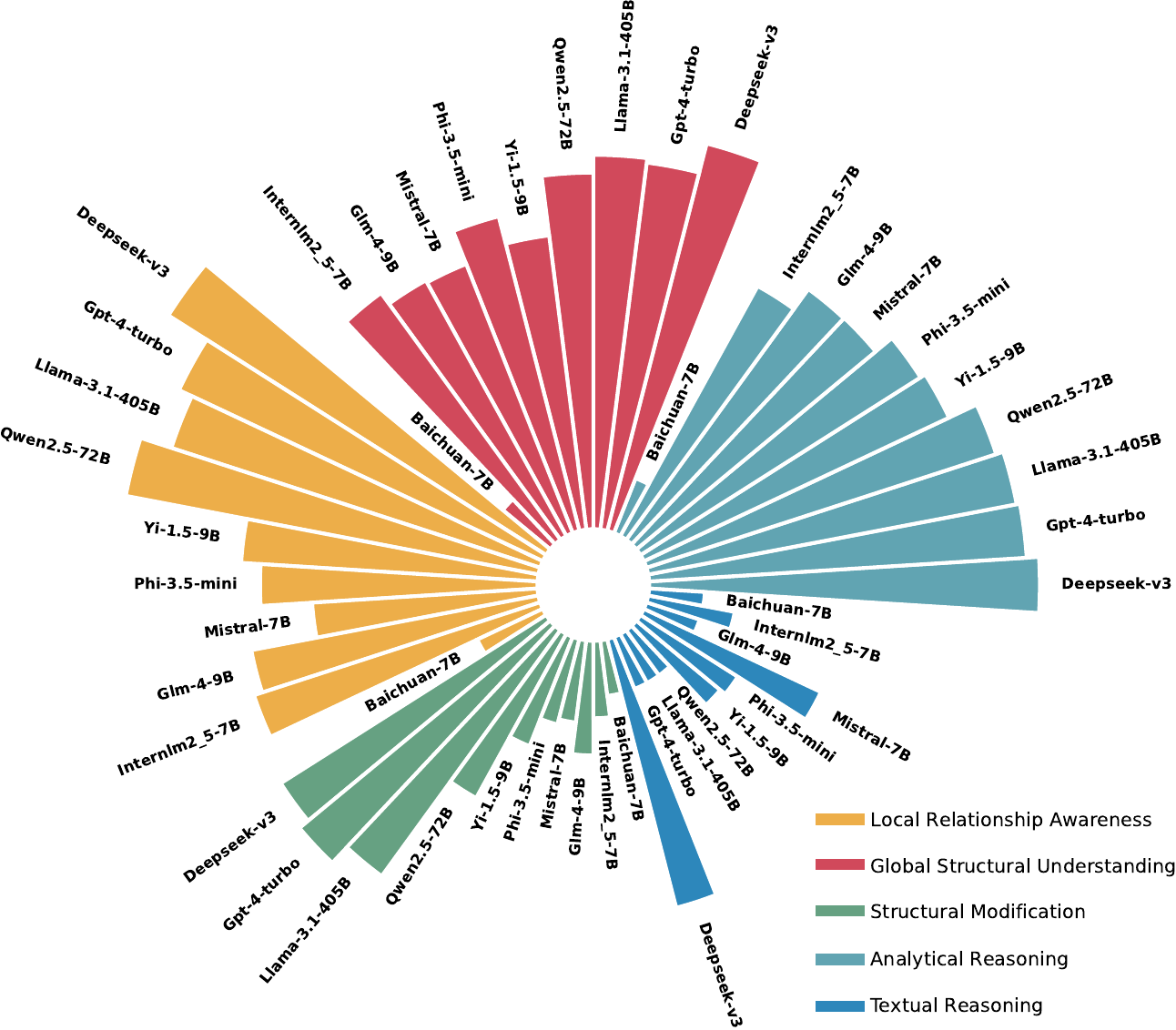}
    \caption{Performance of each Family's State-of-the-art Models over Five Hierarchical Reasoning Capability Dimensions.}
\end{subfigure}
\begin{subfigure}[b]{0.9\linewidth}
    \includegraphics[width=\linewidth]{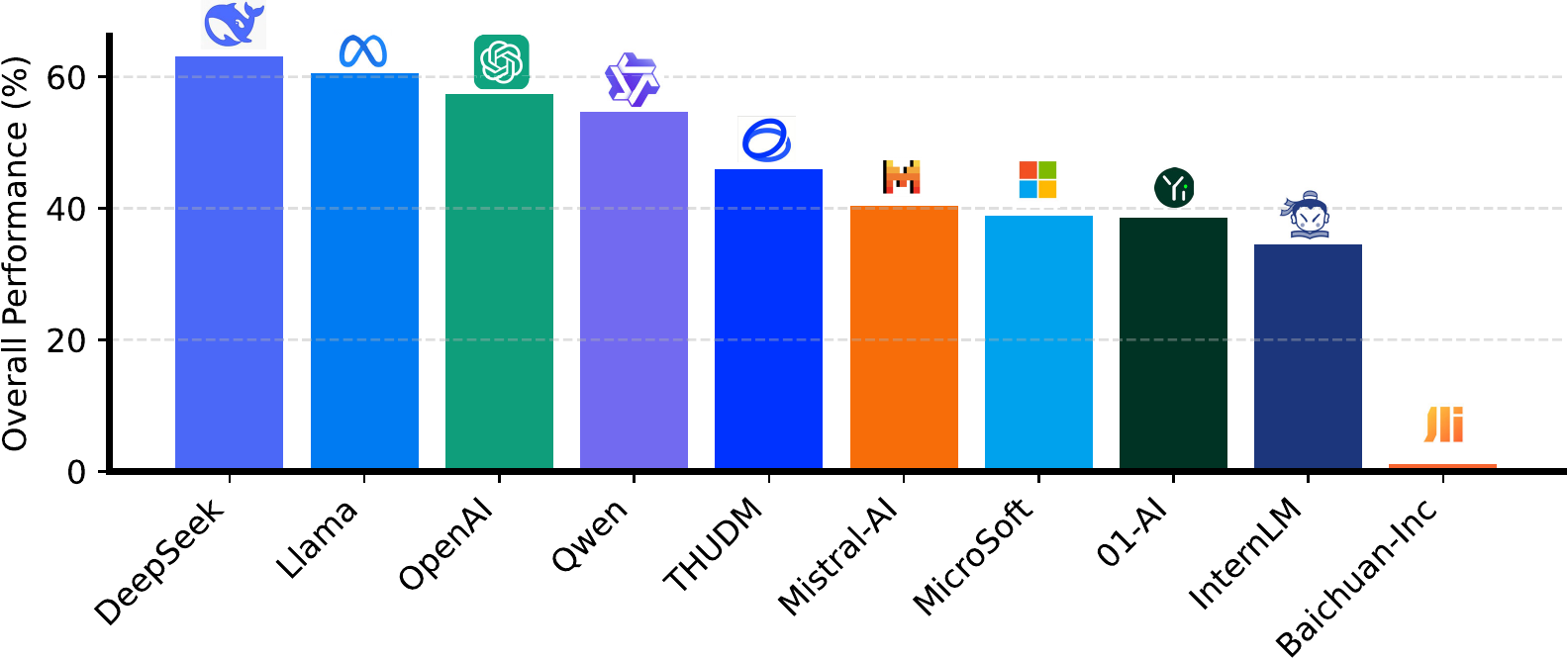}
    \caption{Overall Performance of LLM Families on HiBench.}
\end{subfigure}
\caption{Performance Distribution of LLM Model Families on HiBench.}
\label{fig:radar_chart}
\end{figure}

As diverse task demands drive the evolution of cognitive competencies, HiBench proposes 30 carefully designed tasks with 39,519 queries to assess LLMs' hierarchical reasoning capabilities, which are systematically organized across six distinct scenarios, encompassing both fundamental and practical aspects of hierarchical reasoning.
The fundamental aspect comprises three scenarios: \textit{Binary Tree}, \textit{Multiple Tree}, and \textit{JSON}, accounting for 22 tasks.
These scenarios were selected to assess LLMs' behavior in processing and manipulating abstraction hierarchical structures, which reflect basic yet essential hierarchical reasoning capabilities.
The practical aspect includes three real-world scenarios: \textit{Code}, \textit{Formula}, and \textit{Paper}, consisting of eight specific tasks.
These scenarios embed the complexity within the application context into hierarchical reasoning to assess how well LLMs handle hierarchical information in various practical domains, serving as a robust indicator of advanced hierarchical reasoning capabilities.

\begin{figure*}[t!]
    \centering
    \includegraphics[width=0.95\linewidth]{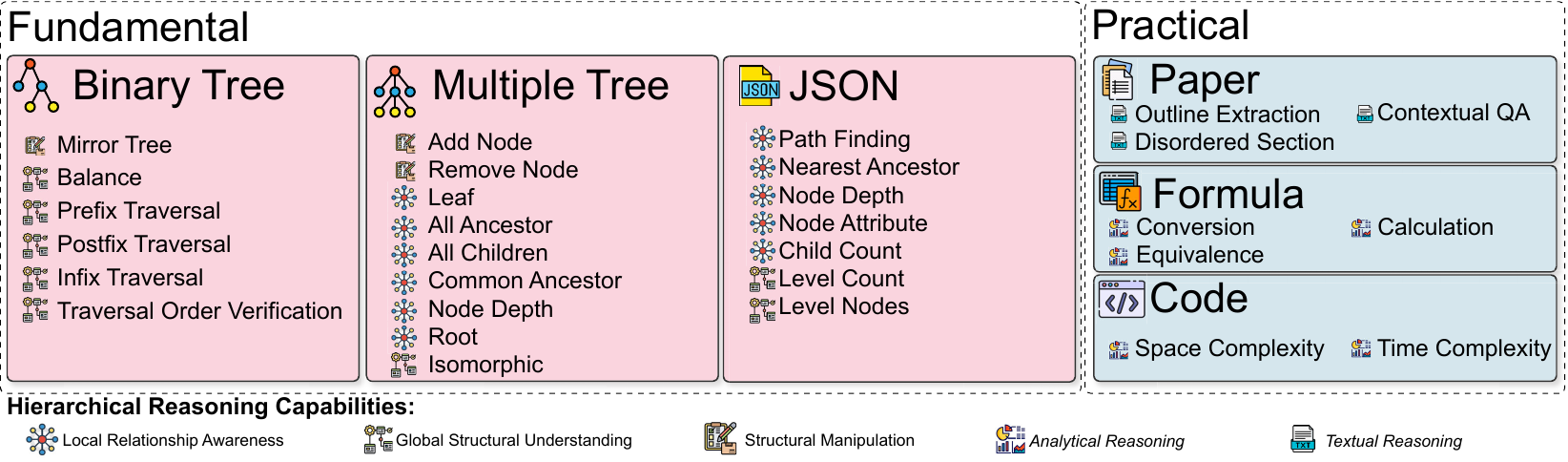}
    \caption{Comprehensive Breakdown of Hierarchical Scenarios Categories and Tasks in HiBench.}
    \label{fig:Tasks}
\end{figure*}

To provide a comprehensive and thorough assessment of LLMs' hierarchical reasoning capabilities, HiBench establishes a multi-dimensional evaluation system comprising five essential dimensions: \textit{Relationship Awareness}, \textit{Structural Understanding}, \textit{Structural Manipulation}, \textit{Analytical Reasoning}, and \textit{Textual Reasoning}.
These dimensions capture distinct yet progressive aspects of hierarchical reasoning, from recognizing and navigating structures to performing complex reasoning tasks.
Furthermore, HiBench incorporates comparative experiments across varying \textit{structure complexity}, \textit{contextual learning paradigms}, and \textit{structure representations}, enabling the investigation of critical factors affecting LLMs' performance across the five evaluation dimensions.
As summarized in Figure~\ref{fig:radar_chart}, extensive experimental results demonstrate that current popular LLMs from various families possess preliminary hierarchical reasoning capabilities with general performance over 40.0\% accuracy.  
Our findings reveal several key insights:
 \textbf{1)} LLMs generally demonstrate more substantial capabilities in relationship awareness, structural understanding, and analytical reasoning than structural manipulation and textual reasoning;
 \textbf{2)} As the complexity of the hierarchical structure increases, whether in depth or breadth, the challenge for LLMs also intensifies;
 \textbf{3)} Structure representation with explicit hierarchical information enhances LLMs' reasoning capabilities;
 \textbf{4)} LLMs perform more effectively when contextual semantics align with real-world hierarchical relationships;
 \textbf{5)} Contextual learning strategies can enhance LLMs' performance, while the benefits of simple chain-of-thought (CoT) prompting are limited.

Since LLMs still have room for improvement in their hierarchical structure reasoning capabilities, we construct a small yet well-designed instruction dataset consisting of 14,623 question-answer pairs across six scenarios based on our findings. 
The dataset was designed to target LLMs' weaknesses in hierarchical reasoning, focusing on complex structures, implicit representations, and counterfactual configurations. 
After instruction fine-tuning, two small-scale LLMs, Llama-3.1-8B \cite{dubey2024llama3} and Qwen2.5-7B \cite{yang2024qwen2_5}, achieved higher performance on HiBench by 88.84\% (Llama-3.1-8B) and 31.38\% (Qwen2.5-7B) across all tasks, compared to their vanilla versions. 
They even exceed large-scale models like LLaMA-3.1-405B by up to 7.31\% (Llama-3.1-8B) and 18.06\% (Qwen2.5-7B), and the GPT-4 \cite{achiam2023gpt4} by up to 6.53\% (Qwen2.5-7B) and 0.2\% (Llama-3.1-8B), which indicates a small-scale high-quality dataset can inspire LLMs' hierarchical reasoning capabilities. 
However, the performance on some tasks remains far less than average, making it an open question to comprehensively boost LLM's hierarchical reasoning ability.

Our contributions are summarized as follows:
\begin{itemize}
    \item We propose HiBench, the first benchmark specifically designed to evaluate hierarchical reasoning capabilities in LLMs.
    \item We conduct extensive experiments evaluating 20 LLMs from 10 families, uncovering both the capabilities and challenges of LLMs in hierarchical reasoning, and providing new insights for further improvement.
    \item By constructing an instruction dataset that targets LLMs' weaknesses in hierarchical reasoning and finetuning small-scale LLMs, we enhance their effectiveness in hierarchical reasoning tasks, outperforming GPT-4 by 6.53\% at most.
\end{itemize}

%% file: 2-related-work.tex
\section{Related Works}
\subsection{LLMs on Structure Reasoning}
With the popularity of LLMs, researchers have begun to deeply explore the combination of these LLMs and structure data~(\emph{e.g.}, graph). Early research mainly focused on empirical performance evaluation. For example, GraphBERT~\cite{zhang2020graphbertattentionneededlearning}, GraphTransformer~\cite{yun2019graph}, and GraphT5~\cite{li2023graphix} are all exploring whether LLM can understand structured graph data. These studies lay a solid foundation for applying LLM in graph data. In addition, Luo~\emph{et al.}~\cite{luo2024graphinstruct} systematically evaluated LLM's capability to perform graph reasoning tasks through GraphInstruct. At the same time, Dai~\emph{et al.}~\cite{dai2024revisiting} pointed out that there is still a significant gap in LLM's understanding of graph structures. However, most of these studies are limited to horizontal graph reasoning tasks, ignoring the importance of evaluating LLM's capability for hierarchical structure reasoning. This limitation prompts us to further explore the application potential of LLM in more complex graph structures.

\subsection{Implicit Hierarchical Thinking in LLMs}

The investigation into \textit{Implicit Hierarchical Thinking} via LLMs has emerged as a prominent research focus in recent years. It was noted that LLMs may not always exhibit a step-by-step reasoning process in implicit reasoning but rather output results directly through their internal hierarchical structure. For example, Deng~\emph{et al.}~\cite{deng2024explicitcotimplicitcot} proposed a transition method from the explicit CoT to implicit CoT to improve the reasoning capability of LLMs through knowledge distillation. Regarding task planning, LLMs can solve complex problems through hierarchical decomposition. For example, Yao~\emph{et al.}\cite{yao2023tree} proposed the Tree of Thoughts framework to optimize LLMs' problem-solving capability through a hierarchical structure. In addition, He~\emph{et al.}\cite{he2024planning} proposed a dual-process framework for dialogue planning that simulates human hierarchical thinking in planning tasks. Research on implicit hierarchical thinking in LLMs has significantly progressed in various areas. However, there is still a need to explore further how to utilize the hierarchical thinking capabilities of LLMs better to enhance their performance in complex tasks.

%% file: 3-hibench.tex
\begin{figure*}[h!]
    \centering
    \includegraphics[width=0.95\linewidth]{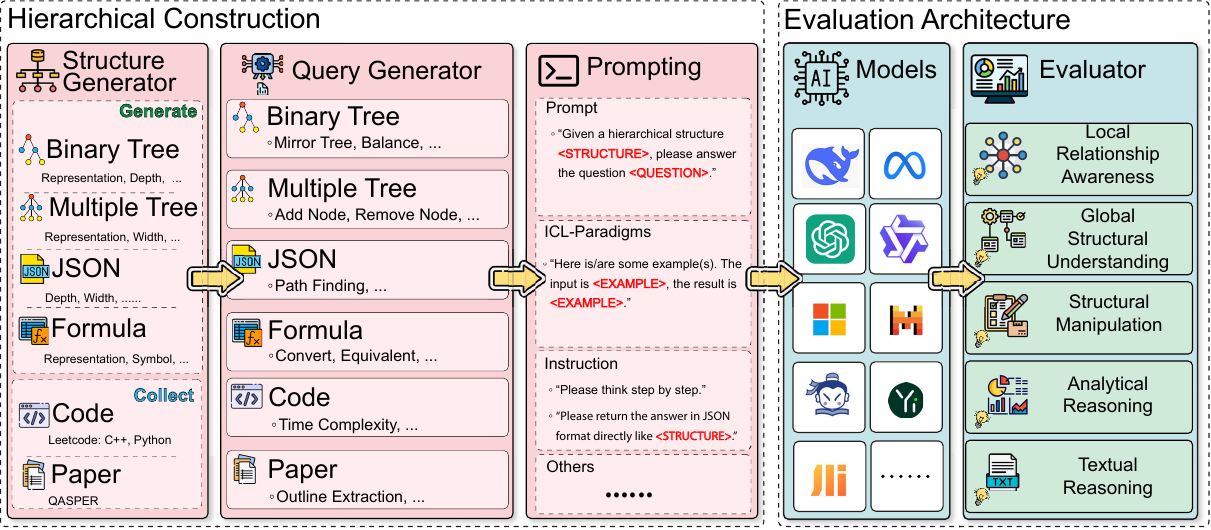}
    \caption{Overview of the Paradigm for HiBench: Hierarchical Construction and Evaluation Architecture.}
    \label{fig:overview}
\end{figure*}

\section{Task Taxonomy}
\label{sec:tasks}

This section provides an overview of task taxonomy in Hibench, illustrated in Figure~\ref{fig:Tasks}, focusing on their design principles.
Since diverse task scenarios demand varying hierarchical reasoning capabilities in LLMs, our taxonomy systematically categorizes tasks based on their complexity and cognitive requirements, encompassing fundamental and practical aspects.
This categorization reveals the core requirement of LLMs for hierarchical reasoning capability and facilitates a precise evaluation of LLMs' abilities across different contexts, offering targeted insights for future model improvements.

\subsection{Fundamental Aspect}

Fundamental tasks focus on assessing LLMs' hierarchical reasoning capability to understand and manipulate structures over hierarchical structure data, such as tree structure and semi-structured JSON files.
These tasks aim to evaluate whether LLMs can effectively perform hierarchical reasoning on diverse hierarchical data representations.

\paragraph{\textbf{Scenario 1: Binary Tree}}  Binary trees are a fundamental data structure widely used in computer science. Using the generalized random structure generator, we constructed the corresponding binary tree with varying difficulty in multiple dimensions of depth, then designed six hierarchical tasks, \textit{Balance, Prefix Traversal, Infix Traversal, Postfix Traversal, Traversal Order Verification} and \textit{Mirror Tree} to evaluate LLM's hierarchical understanding of tree structures.
    
\paragraph{\textbf{Scenario 2: Multiple Tree}} Multiple trees are widely utilized in diverse domains such as database indexing, file systems, and network architectures, where they excel in representing complex hierarchical arrangements that surpass the limitations of binary trees. Therefore, we use a randomized tree generator to construct multinomial trees of varying difficulty in multiple dimensions of breadth and depth, then developed nine hierarchical tasks: \textit{Add Node, All Ancestor, All Children, Common Ancestor, Isomorphic, Remove Node, Node Depth, Leaf} and \textit{Root}. These tasks are designed to fully evaluate the hierarchical reasoning capabilities of LLMs.

\paragraph{\textbf{Scenario 3: JSON}} JSON is a semi-structured data interchange format that can represent hierarchical data structures using nested objects and arrays. As a standardized format, JSON is widely supported and understood. Therefore, we randomly generated JSON files with different breadth and depth and designed the following seven tasks: \textit{Child Count, Node Depth, Level Count, Node Attribute, Level Nodes, Path Finding} and \textit{Nearest Ancestor}. These tasks examine the LLM's capability to parse JSON data, understand hierarchies, and analyze paths in multiple dimensions.
\subsection{Practical Aspect}
In contrast to the fundamental aspect, the practical aspect emphasizes real-world applications ranging in information format and structural diversity, such as formulas, code, and academic papers. 
These tasks assess the versatility and capability of LLMs to apply hierarchical reasoning in addressing noisy and complex real-world challenges.

\paragraph{\textbf{Scenario 4: Formula}} Mathematical formulas can exhibit certain hierarchical information in prefix, infix, and postfix notations. Therefore, we randomly generated a series of formulas of varying difficulty using formula length, number size, and symbolic complexity as variables and constructed three types of formula comprehension tasks: 
\textit{Conversion, Calculation} and \textit{Equivalence}, to assess LLM's formula hierarchical understanding capability.

\paragraph{\textbf{Scenario 5: Code}} The structure and logic of the code are hierarchical. Therefore, we collected different types of C++ and Python code on the LeetCode platform and constructed two types of tasks, \textit{Space Complexity} and \textit{Time Complexity}, aiming at evaluating the capability of the LLM to understand and analyze the code.

\paragraph{\textbf{Scenario 6: Paper}} Papers are hierarchically structured textual information, so we carefully screened several papers as datasets based on the QASPER \cite{dasigi2021dataset} dataset and assessed LLM's hierarchical reasoning capabilities on academic papers based on three types of tasks \textit{Contextual QA}, \textit{Disordered Section} and \textit{Outline Extraction}.

\section{The HiBench}

In this section, we introduce the overall architecture of HiBench, a comprehensive and systematic benchmark developed to assess the hierarchical reasoning capabilities of LLMs, thereby facilitating user adoption and utilization by ensuring a well-established workflow.
The architecture is organized into two main components: \textit{Hierarchical Dataset Constructor} and \textit{Evaluator}.
\textit{Hierarchical Dataset Constructor} systematically generates benchmark data with varying complexity.
\textit{Evaluator} quantifies model performance from five refined capability dimensions.
An overview of this architecture is illustrated in Figure~\ref{fig:overview}, with more details provided in Appendix~\ref{app:codebase}.

\subsection{Hierarchical Dataset Generator}
In this section, the generation process of HiBench is introduced, consisting of three main stages: \textit{Structure Generator}, \textit{Query-Answer Generation}, and \textit{Prompting Production}.
Firstly, building upon the previously mentioned task taxonomy in Section~\ref{sec:tasks}, the difficulty of specific tasks within each scenario varies due to the intricacy of their hierarchical structures.
To ensure a thorough evaluation, \textit{Structure Generator} constructs various hierarchical structures for subsequence processes across different scenarios.
Secondly, \textit{Query-Answer Generation} utilizes these pre-constructed structures to generate corresponding queries for sub-tasks in different scenarios.
Finally, \textit{Prompting Production} refines these queries into well-structured prompts, adapting them for LLMs and ensuring consistency across evaluations.

\paragraph{\textbf{Structure Generator.}}
Given that the task taxonomy of HiBench encompasses diverse application scenarios characterized by specialized requirements and varying levels of complexity, tailoring structure generators is necessary. 
For the four scenarios, Binary Tree, Multiple Tree, JSON, and Formula, their structure generators autonomously and efficiently construct large-scale, diverse structure instances, providing rich and challenging hierarchical structures to support comprehensive evaluation.
Specifically, the tree generator constructs hierarchical tree data based on key features such as out-degree, depth, and node number.
The JSON generator constructs data by considering two key dimensions: width and depth, while the formula generator constructs data based on numerical complexity, symbolic complexity, and length of formulas to ensure varying degrees of complexity.
In contrast, for the other two scenarios, Code and Paper, their structure generators collect a large amount of hierarchically structured data from existing real-world sources as initial data.
Specifically, the code generator collects code samples from GitHub\footnote{https://github.com} and LeetCode\footnote{https://leetcode.com/}, while the paper generator reorganizes paper data from QASPER~\cite{dasigi2021dataset}. These data encompass the hierarchical structures of codes and papers, ensuring their diversity and real-world representativeness.

\paragraph{\textbf{Query-Answer Generator.}}
Since each task scenario consists of multiple sub-tasks, we generate corresponding query-answer pairs based on the structures constructed in the previous section. Various algorithms are employed to tailor queries to specific sub-tasks. For instance, the prefix traversal algorithm generates query-answer pairs for the corresponding sub-task based on the binary tree structure. Notably, we use manual annotations to generate query-answer pairs for paper scenarios without fixed answers.

\paragraph{\textbf{Prompting Producer.}}
To optimize query-answer pairs for LLM-generated responses, we develop a set of carefully crafted prompts, including In-context Learning (ICL)~\cite{dong2024survey}, and CoT~\cite{zhang2023automatic}, enabling comprehensive evaluation across different requirements. We also provide instructions to restrict the output format, ensuring responses align with specific structural or stylistic guidelines.

\begin{table}[!t]
\centering
\caption{Basic Statistics of HiBench.}
\label{tab:basic-stat}
\resizebox{\linewidth}{!}{
\begin{tabular}{ccccc}
\toprule
\textbf{Scenarios}       & \textbf{Tasks} & \textbf{Sub-Tasks} & \textbf{Queries} & \textbf{Avg. Length} \\ 
\midrule
Binary Tree   & 6     & 216      & 4,968    & 2,824.3      \\
Multiple Tree & 9     & 162      & 9,558    & 915.0       \\
JSON          & 7    & 300      & 2,586    & 1,349.8      \\
Formula       & 3     & 1,458     & 18,954   & 516.8       \\
Code          & 2     & 12       & 1,200    & 1,296.9      \\
Paper         & 3     & 9        & 2,253    & 26,759.2     \\ 
\midrule
\textbf{HiBench}           & 30    & 2,157     & 39,519  & 1,725.9      \\ 
\bottomrule
\end{tabular}
}
\end{table}
Table~\ref{tab:basic-stat} provides a detailed statistical summary of HiBench across various scenarios, comprising 30 tasks and 39,519 queries. Our benchmark dataset offers a robust foundation for evaluating the hierarchical reasoning capabilities of LLMs across diverse queries.

\subsection{Evaluator}

\input{tabels/main-result}
Hierarchical reasoning in human cognition progresses from understanding local relationships to grasping global structures, performing structural operations, and reasoning within increasingly complex information contexts, reflecting a growing demand for advanced reasoning capabilities.

Building on this progression, we define five key dimensions to evaluate LLMs' hierarchical reasoning capabilities, mirroring the stages of human cognition progress in hierarchical reasoning:
\begin{itemize}
    \item \textit{Local Relationship Awareness} assesses the capability to recognize immediate connections, such as parent-child relationships in trees or code dependencies.
    \item \textit{Global Structural Understanding} extends this to comprehending the overall organization of a structure.
    \item \textit{ Structural Manipulation} evaluates the capability to modify structures, such as code refactoring or tree transformations.
    \item \textit{Analytical Reasoning} measures the capability to derive insights through logical and quantitative analysis.
    \item \textit{Textual Reasoning} examines the capacity to generate coherent and contextually appropriate descriptions of hierarchical structures.
\end{itemize}

Based on the characteristics of each task, we categorize LLMs' performance into these five evaluation dimensions, as illustrated in Figure~\ref{fig:Tasks}. It aligns tasks with the cognitive and computational demands of structured data processing, forming an integral framework for assessing hierarchical reasoning capabilities. 

%% file: tabels/main-result.tex
\begin{table*}[!h]
\centering
\caption{HiBench Leaderboard: Categorizing Models by Open-Source Status and Family.$^*$}
\label{Tab:LLMPerformance}
\begin{tabular}{clcccccccc}
\toprule
\multirow{2}{*}{{Model Family}} & \multirow{2}{*}{{Model Name}} & \multicolumn{3}{c}{{{\raisebox{-0.14\height}{\includegraphics[height=1.1em]{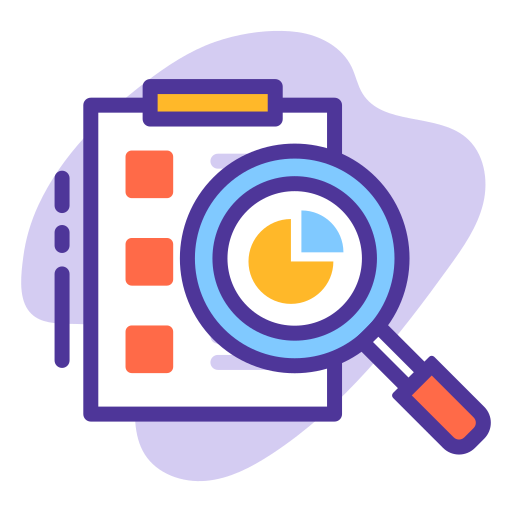}}}\hspace{0.2em}{Fundamental Aspect}}} & \multicolumn{3}{c}{
{{\raisebox{-0.14\height}{\includegraphics[height=1.1em]{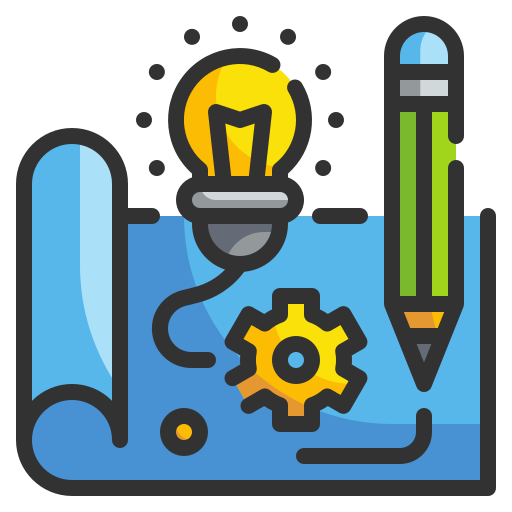}}}\hspace{0.2em}{Practical Aspect}}
} & \multirow{2}{*}{\textbf{Average}} & \multirow{2}{*}{\textbf{Rank}}  \\
\cmidrule(lr){3-5} 
\cmidrule(lr){6-8} 
& & {{\raisebox{-0.14\height}{\includegraphics[height=1.1em]{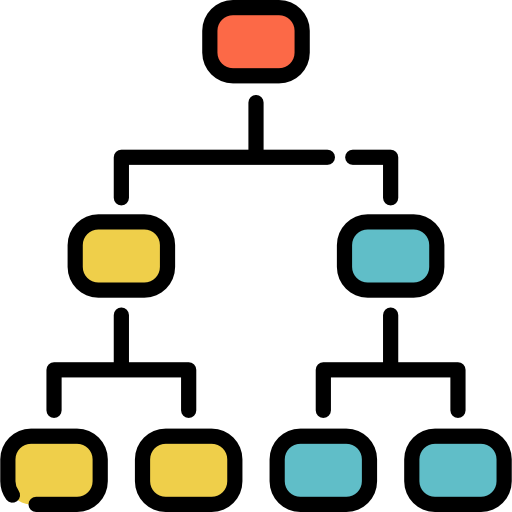}}}\hspace{0.2em}\textit{{Binary}}} & {{\raisebox{-0.14\height}{\includegraphics[height=1.1em]{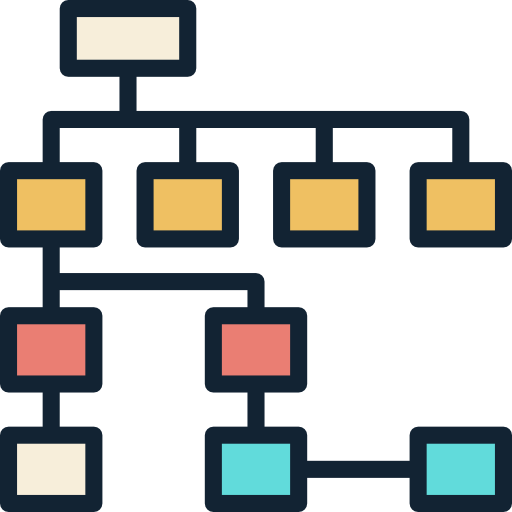}}}\hspace{0.2em}\textit{{Multiple}}} & {{\raisebox{-0.14\height}{\includegraphics[height=1.1em]{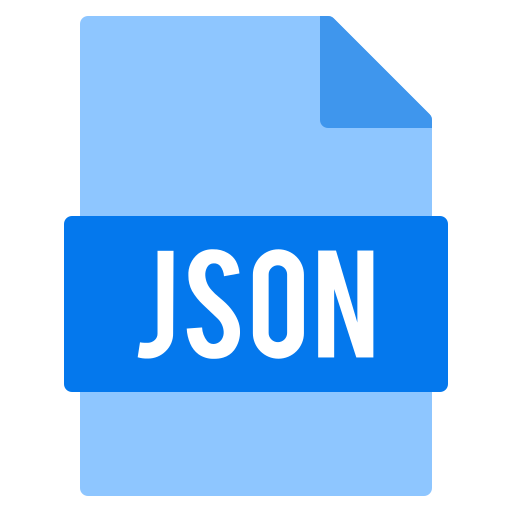}}}\hspace{0.2em}\textit{{JSON}}}  & {{\raisebox{-0.14\height}{\includegraphics[height=1.1em]{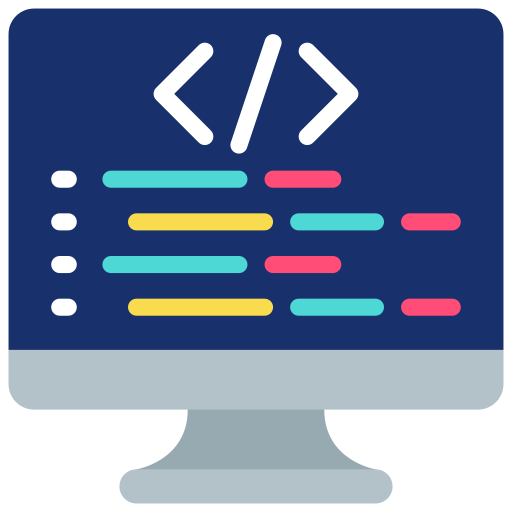}}}\hspace{0.2em}\textit{{Code}}} & {{\raisebox{-0.14\height}{\includegraphics[height=1.1em]{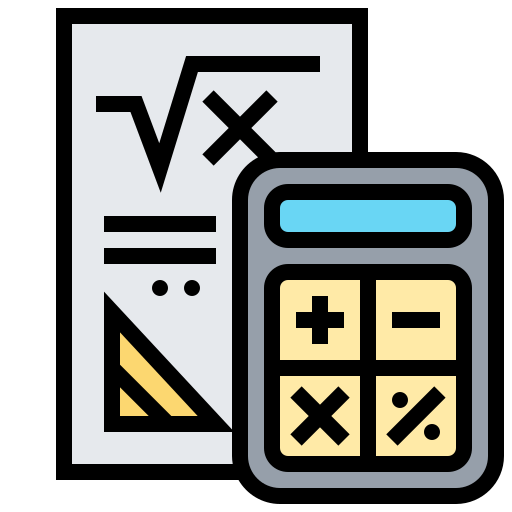}}}\hspace{0.2em}\textit{{Formula}}} & {{\raisebox{-0.14\height}{\includegraphics[height=1.1em]{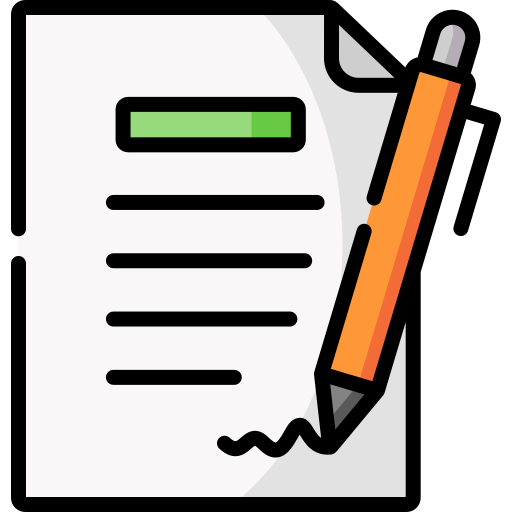}}}\hspace{0.2em}\textit{{Paper}}}\\
\midrule
\rowcolor{red!20} \multicolumn{10}{c}{\textit{\textbf{Closed-Sourced}}} \\
\midrule
\multirow{2}{*}{OpenAI} & GPT-3.5 \cite{openai2022chatgpt} & 39.19	&	54.22	&	53.49	&	66.50	&	\underline{54.54}	&	-$^{**}$	&	53.59	&	6 \\
                         & GPT-4 \cite{achiam2023gpt4}          & 56.29	&	\underline{73.64}	&	63.59	&	\underline{70.75}	&	54.37	&	38.35	&	\underline{59.50}	&	2 \\
                         
\midrule
\rowcolor{blue!20} \multicolumn{10}{c}{\textit{\textbf{Open-Sourced}}} \\
\midrule
\multirow{1}{*}{01-AI} & Yi-1.5-9B \cite{young2024yi}   &     26.18	&	47.67	&	54.23	&	66.75	&	42.42	&	10.27	&	41.25	&	9 \\
\midrule
\multirow{6}{*}{Qwen} & Qwen2.5-0.5B \cite{yang2024qwen2_5}  & 1.99	&	15.03	&	15.31	&	2.50	&	8.83	&	3.98	&	7.94	&	19 \\
                               & Qwen2.5-1.5B \cite{yang2024qwen2_5}  & 19.43	&	43.30	&	35.47	&	56.75	&	21.17	&	25.54	&	32.02	&	17 \\
                               & Qwen2.5-3B  \cite{yang2024qwen2_5}  & 24.32	&	48.95	&	39.71	&	60.25	&	39.78	&	32.38	&	40.52	&	10 \\
                               & Qwen2.5-7B  \cite{yang2024qwen2_5}  & 33.11	&	59.01	&	48.58	&	69.25	&	41.27	&	42.08	&	49.93	&	7 \\
                               & QwQ-32B \cite{qwq-32b-preview}  & 42.91	&	\textbf{74.12} &	\underline{73.80}	&	13.75	&	16.82	&	17.22	&	39.77	&	12 \\
                               & Qwen2.5-72B \cite{yang2024qwen2_5} & 45.61	&	63.08	&	70.50	&	69.25	&	51.75	&	39.72	&	56.65	&	3\\
\midrule
\multirow{1}{*}{Baichuan Inc.} & Baichuan-7B \cite{yang2023baichuan}  & 2.62	&	1.75	&	4.33	&	0.25	&	0.93	&	0.00	&	1.65	&	20 \\
\midrule
\multirow{1}{*}{DeepSeek} & DeepSeek-V3 \cite{liu2024deepseek-v3}  &   \textbf{70.23}	&	72.59	&	\textbf{77.65}	&	\textbf{72.50} &	\textbf{57.40}	&	\underline{42.99}	&	\textbf{65.56}	&	1 \\
\midrule
\multirow{5}{*}{Meta}  & Llama-3.2-1B \cite{dubey2024llama3} & 9.57	&	20.45	&	23.18	&	44.25	&	17.02	&	13.30	&	21.30	&	18 \\
                       & Llama-3.2-3B \cite{dubey2024llama3} & 31.79	&	34.70	&	29.31	&	60.50	&	23.20	&	28.07	&	34.59	&	15 \\
                       & Llama-3.1-8B \cite{dubey2024llama3} & 14.74	&	28.76	&	49.29	&	61.75	&	26.35	&	16.23	&	32.85	&	16 \\
                       & Llama-3.1-70B \cite{dubey2024llama3} & 54.73	&	65.04	&	63.03	&	68.75	&	42.14	&	\textbf{43.71}	&	56.24	&	4 \\
                       & Llama-3.1-405B \cite{dubey2024llama3} & \underline{64.49}	&	59.95	&	55.61	&	53.37	&	63.45	&	16.46	&	55.53	&	5 \\                
\midrule
\multirow{1}{*}{Microsoft} & Phi-3.5-mini-3.8B \cite{abdin2024phi3} & 27.36	&	38.24	&	57.52	&	65.25	&	41.76	&	11.51	&	40.27	&	11 \\
\midrule
\multirow{1}{*}{Mistral} & Mistral-7B \cite{jiang2023mistral7b} & 26.14	&	31.00	&	49.19	&	58.75	&	40.08	&	27.10	&	38.71	&	13 \\

\midrule
\multirow{1}{*}{SHAILib} & InternLM2.5-7B \cite{cai2024internlm2} & 34.28	&	39.02	&	54.62	&	65.75	&	23.27	&	6.24	&	37.20	&	14 \\
\midrule
\multirow{1}{*}{THUDM} & GLM-4-9B \cite{glm2024chatglm} & 36.72	&	46.26	&	48.60	&	66.50	&	32.11	&	37.60	&	44.63	&	8 \\
\midrule
\multicolumn{2}{c}{\textbf{Average Performance}}        & 33.08    &	44.61	&	49.94	&	55.53	&	34.43	&	23.83 & 40.48 & -\\
\bottomrule
\end{tabular}

\raggedright\footnotesize{\hspace{0.2cm}$^*$ The best results for each category are marked in \textbf{bold}, and the second-best results are marked with \underline{underline}.\\
\hspace{0.2cm}$^{**}$ Queries exceed the maximum context token limitation of GPT-3.5.}
\end{table*}

%% file: 5-results.tex
\section{Experimental Results}
In this section, we present the experimental results evaluating the performance of LLMs on hierarchical reasoning using our HiBench. We use \textit{Accuracy} as the primary metric to assess LLM performance, defined as $Accuracy = \frac{\#Correct}{\#Total}$.

\subsection{LLMs' Performance on HiBench}
\paragraph{Overview.} Table~\ref{Tab:LLMPerformance} presents the performance of the 20 most popular and powerful LLMs across tasks in our HiBench categorized by model families and scenarios. Current LLMs achieve an average score of 40.48\% over our HiBench, demonstrating the general level of hierarchical reasoning capabilities. Specifically, DeepSeek-V3 \cite{liu2024deepseek-v3} stands out with the highest performance, achieving a mean score of 65.56\%. It surpasses the benchmark average by 62.0\% and outperforms the most powerful closed-sourced LLM, GPT-4 \cite{achiam2023gpt4}, by 10.2\%. Although many LLMs demonstrate basic hierarchical reasoning capabilities in managing Multiple Tree, JSON, and Code scenarios, these LLMs still face challenges in more complex scenarios, especially in the practical aspects. These findings highlight the importance of continued model refinement to enhance LLMs' behavior on more advanced and nuanced hierarchical tasks.

\begin{figure*}[h!]
    \centering
    \includegraphics[width=0.83\linewidth]{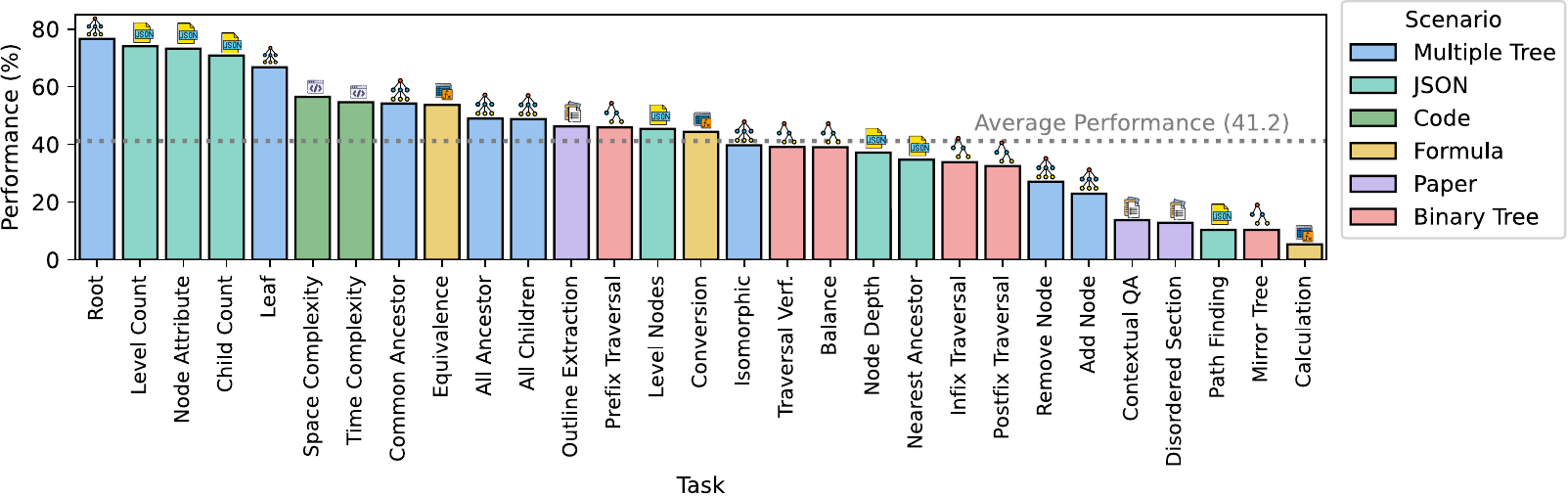}
    \caption{Average Performance of LLM Tasks across Capability Dimensions and Scenarios.}
    \label{fig:task-performance}
\end{figure*}

\paragraph{Scenario Performance.} Figure~\ref{fig:task-performance} presents the performance of LLMs across various hierarchical reasoning tasks, categorized by different scenarios such as Multiple Tree, JSON, Code, Formula, Paper, and Binary Tree. The results show that LLMs perform well on tasks such as \textit{Root Identification}, \textit{Level Counting}, and \textit{Node Attribute Recognition}, with performance values reaching up to 70\%. These tasks, which require minimal reasoning complexity and exhibit clear structural patterns, allow LLMs to achieve high accuracy. In contrast, tasks like \textit{Path Finding}, \textit{Mirror Tree}, and \textit{Calculation} are generally below 30\% and pose significant challenges due to their requirements for multi-step reasoning, structural transformations or numerical computations. The average performance across all tasks is 41.2\%.

\paragraph{Capability Dimensions.} As illustrated in Figure~\ref{fig:radar_chart}~(a), LLMs perform well in local relationship awareness, global structural understanding, and analytical reasoning but struggle with structural modifications and textual reasoning. DeepSeek-V3 leads overall, scoring 75.96 in local relationship awareness, 66.77 in global structural understanding, and 64.95 in analytical reasoning, yet still lags in structural modifications and textual reasoning. Similarly, GPT-4 and Qwen2.5-72B-Instruct maintain strong local relationship awareness and analytical reasoning scores but steeply decline in structural modifications. Smaller models, including Yi-1.5-9B and Phi-3.5-mini-3.8B, perform poorly in structural modifications and TR. Baichuan-7b fails in all dimensions, and we conduct case studies in Appendix \ref{app:casestudy-baichuan} to discuss its failure. The performance disparity underscores that while LLMs recognize and analyze hierarchical structures well, they struggle with modifying and reasoning through hierarchical text, highlighting further gains in these tasks.

\begin{figure}[!h]
    \centering
    \includegraphics[width=\linewidth]{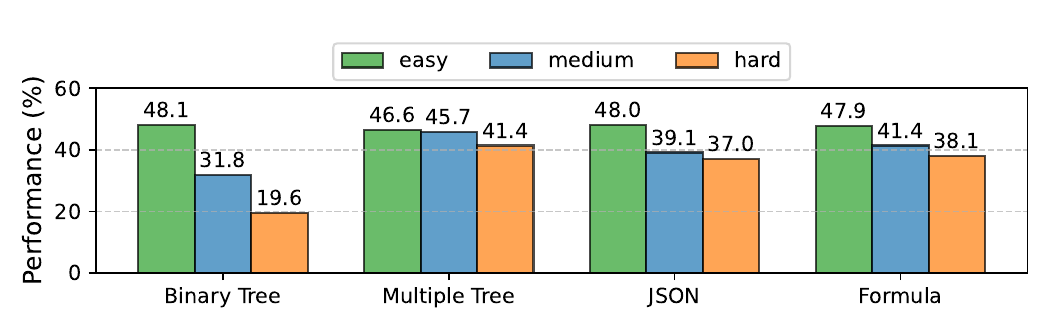}
    \caption{Impact of Structural Complexity on LLM Hierarchical Reasoning Capabilities.}
    \label{fig:structure-complexity}
\end{figure}

\paragraph{Structure Complexity Impact.} Figure~\ref{fig:structure-complexity} presents the performance of LLMs across varying structure complexity levels. The results demonstrate that LLMs perform significantly better on simpler structures, with accuracy peaking for the \textit{easy} structures in all scenarios. In contrast, as structure complexity increases, performance declines noticeably, with the \textit{hard} structures yielding only 19.6\% accuracy in the Binary Tree scenario and 37.0\% in the JSON scenario. These findings suggest that while LLMs can effectively handle tasks with simpler structures and clear patterns, they struggle with more complex hierarchical structures, highlighting the need for further model development to address more challenging hierarchical structures.

\subsection{Insightful Findings}
\subsubsection{Structure Representation affect LLM Reasoning}
As the structure representation mode can affect an LLM's capability to capture hierarchical information, we explore the impact of different representation modes on hierarchical reasoning performance. Specifically, we compare two representation modes: \textit{edge} representation and \textit{text tree} representation, over binary and multiple tree tasks. The edge representation describes the hierarchical structure through a list of directed edges, while the text tree representation uses structure symbols to depict the structure. As shown in Figure~\ref{fig:input-mode}~(a), LLMs with text tree representation achieve better performance than those with edge representation on both binary tree and multiple tree tasks, with performance values of 39.6\% and 38.4\%, respectively, indicating that the input representation can affect the LLM's hierarchical reasoning capability. As shown in Figure~\ref{fig:Task1_repre}, the text tree representation provides global hierarchical information, while the edge representation focuses only on local connections, which may limit its effectiveness in capturing the overall structure. It suggests that the text tree representation is more beneficial for tasks requiring a global understanding of the hierarchy.

\subsubsection{Expression of Explicit Structures Improves LLM Comprehension}
\begin{figure}[h!]
    \centering
    \includegraphics[width=\linewidth]{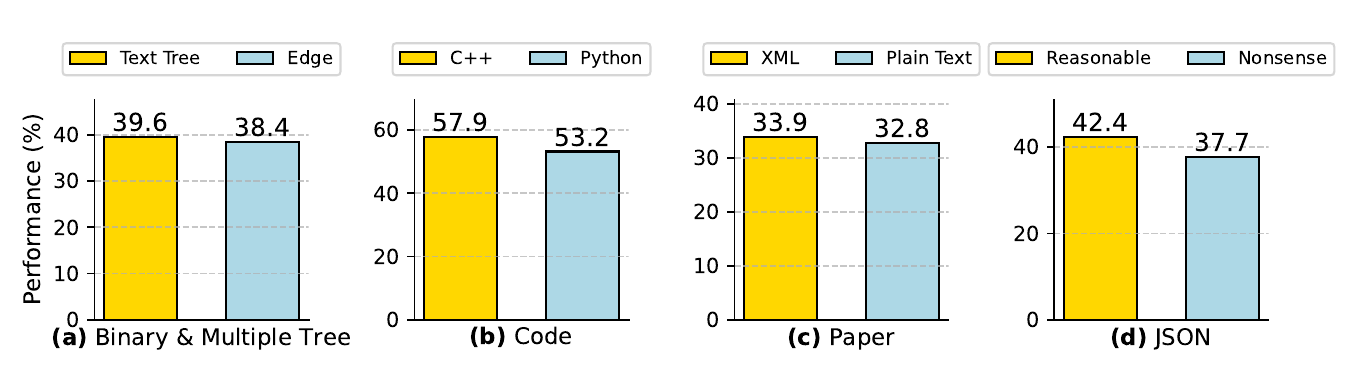}
    \caption{Impact of Input Mode on LLM Hierarchical Reasoning Capabilities.}
    \label{fig:input-mode}
\end{figure}

In practical aspects, such as Code and Paper scenarios, hierarchical structures may present with either implicit or explicit structural constraints. 
In the Code scenario, C++ utilizes explicit structural constraints where double braces mark each block, while Python relies on indentation to represent hierarchical relationships implicitly. 
In the Paper scenario, papers formatted in an XML-like style \cite{che2024hierarchical} offer clear structural indicators, whereas those based on plain text lack explicit hierarchical cues. 
Figure~\ref{fig:input-mode}~(b) and (c) investigate how these kinds of structural constraints impact LLMs' hierarchical reasoning capabilities. It demonstrates that inputs with explicit structural constraints consistently outperform those with implicit or no structure. Specifically, C++ outperforms Python by 4.7\% in the Code scenario, while structured inputs surpass unstructured ones by 1.1\% in the Paper scenario. The superior performance can be attributed to the clarity and consistency of explicit structural formats, such as C++'s strict syntactic rules and XML's hierarchical markers, which define clear boundaries and reduce ambiguity, allowing LLMs to capture the hierarchical information more effectively. It highlights the critical role of explicit hierarchical structural constraints in enhancing the hierarchical reasoning capabilities of LLMs in practical aspects.

\subsubsection{Semantics Enhances LLM Hierarchical Structure Reasoning}
To investigate whether LLMs have captured essential hierarchical reasoning capability from real-world corpora during pre-training, we compare their performance on real-world hierarchical structures and nonsense structures in the JSON scenario. The original JSON scenario is based on real-world hierarchical structures containing semantically meaningful values. We construct nonsense structures involving hierarchies with randomly shuffled labels or meaningless values for comparison. As shown in Figure~\ref{fig:input-mode}~(d), LLMs achieve a significantly higher accuracy of 42.4\% with real-world semantic structures compared to 37.7\% with nonsense structures. This improvement suggests that LLMs have likely learned basic hierarchical structure from real-world data, enabling them to better parse and reason over nested JSON hierarchies.

\subsubsection{Impact of CoT}
CoT is known to improve performance in various tasks by prompting models to break down problems into step-by-step logical sequences. To evaluate its impact on LLM hierarchical reasoning, we conducted experiments of Qwen2.5-7B \cite{yang2024qwen2_5} and QwQ-32B \cite{qwq-32b-preview} on Binary Tree and Multiple Tree scenarios. As shown in Table~\ref{tab:cot_comparison}, CoT slightly improved Qwen2.5-7 B's accuracy in the Multiple Tree task by 3.33\% but had minimal degradation on the Binary Tree task. However, for QwQ-32B, CoT led to a 0.9\% accuracy decrease in Binary Tree and a substantial 31.26\% drop in Multiple Tree performance. These results indicate that CoT does not consistently enhance hierarchical reasoning and may introduce unnecessary steps that complicate intermediate reasoning, as further analyzed in our case study (Appendix~\ref{app:casestudy-cot}). 

\begin{table}[h]
    \centering
    \caption{Performance Comparison of Models Using CoT Reasoning Versus Standard One.}
    \begin{tabular}{lcc}
        \toprule
        & \textbf{Binary Tree} & \textbf{Multiple Tree} \\
        \midrule
        Qwen2.5-7B $w/$ CoT   & 33.11 & 54.86 \\
        Qwen2.5-7B $w/o$ CoT  & 33.06 & 58.19 \\
        \midrule
        $\Delta_{CoT}$ & \cellcolor{lightred} -0.05 $\downarrow$ & \cellcolor{lightgreen} 3.33 $\uparrow$ \\
        \midrule
        QwQ-32B $w/$ CoT   & 42.91 & 74.12 \\
        QwQ-32B $w/o$ CoT  & 42.01 & 42.86 \\
        \midrule
        $\Delta_{CoT}$ & \cellcolor{lightred} -0.9 $\downarrow$ & \cellcolor{lightred} -31.26 $\downarrow$ \\
        \bottomrule
    \end{tabular}
    \label{tab:cot_comparison}
\end{table}

\subsection{Potential Improvement}
Given the nascent state of LLMs' performance in hierarchical reasoning, we conducted a series of experiments, including ICL and instruction fine-tuning, to assess the effectiveness of these methods in improving their hierarchical reasoning capabilities.

\begin{table}[t]
\caption{Instruction Finetuning Performance on HiBench.}
\label{Tab:finetune}
\begin{tabular}{ccccc}
\toprule
\multicolumn{1}{c}{\multirow{2}{*}{Aspects}} & \multicolumn{2}{c}{Qwen2.5-7B} & \multicolumn{2}{c}{Llama-3.1-8B} \\ \cmidrule(lr){2-3} \cmidrule(lr){4-5} 
\multicolumn{1}{l}{}                      & Vanilla        & Finetuned       & Vanilla         & Finetuned        \\ 
\midrule
Fundamental                            & 42.32	& 65.74	& 25.89	& 63.59           \\ 
Practical                             & 54.17	& 61.05	& 37.20	& 55.60           \\
\midrule
\textbf{Average}                                      & 48.25	& \textbf{63.39}	& 31.55	& \textbf{59.59}          \\ 
\bottomrule
\end{tabular}
\end{table}

\begin{figure}[t]
    \centering
    \includegraphics[width=\linewidth]{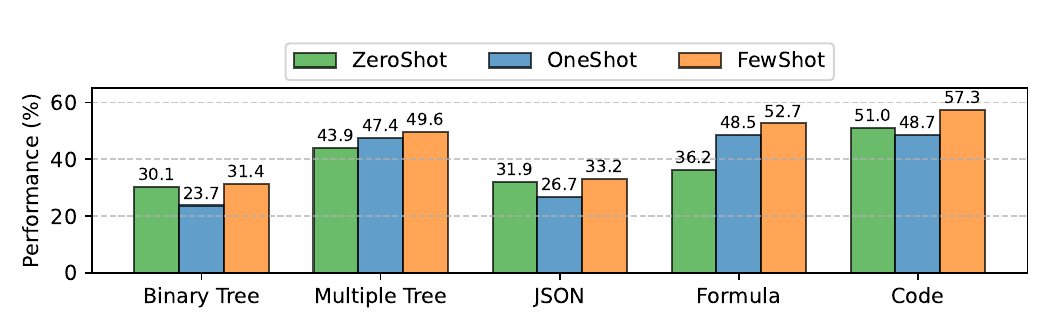}
    \caption{Impact of ICL on Hierarchical Reasoning.}
    \label{fig:example-type}
\end{figure}

Figure~\ref{fig:example-type} demonstrates ICL's impact on LLMs over our HiBench, where zero-shot learning results in the lowest performance, one-shot learning shows moderate improvement and few-shot learning significantly enhances LLMs' hierarchical reasoning capabilities. This trend demonstrates that LLMs benefit from additional context through ICL and that more examples lead to substantially better performance across hierarchical reasoning tasks. However, in some scenarios, such as Binary Tree, JSON, and Code, one-shot learning performs poorly compared to zero-shot. Our case study (Appendix~\ref{app:casestudy-icl}) suggests that LLMs may develop an unintended inductive bias when only a single example is provided, leading to incorrect generalization. This effect is particularly pronounced when the example contains high-frequency patterns or structural repetitions, causing the model to overfit spurious correlations rather than accurately grasping the hierarchical relationships. It highlights the importance of instance diversity in few-shot settings to mitigate biases and enhance LLMs' hierarchical reasoning capabilities.

To further enhance the hierarchical reasoning capabilities of LLMs, we conducted instruction fine-tuning on the Llama-3.1-8B and Qwen2.5-7B models using a well-designed instruction dataset with 14,623 examples targetted to hierarchical reasoning. The dataset focuses more on scenarios with complex hierarchical structures, implicit structural representations, and counterfactual hierarchical configurations. This design aims to enhance LLMs' reasoning across the weaker aspects. As shown in Table~\ref{Tab:finetune}, fine-tuning led to significant performance improvements across fundamental and practical aspects. Specifically, Llama-3.1-8B improved from 31.55\% to 59.59\% on average, and Qwen2.5-7B improved from 48.25\% to 63.39\%. The fine-tuning results highlight the importance of finetuning LLMs to enhance hierarchical reasoning, especially for complex tasks.

%% file: 6-conclusion.tex
\section{Conclusion}
In this paper, we propose HiBench, the first benchmark dedicated explicitly to evaluating the hierarchical reasoning capabilities of LLMs. HiBench spans two key aspects, six scenarios, and 30 tasks, comprising 39,159 queries. Experimental results demonstrate that while existing LLMs show proficiency in basic hierarchical reasoning tasks, they still struggle with more complex hierarchical challenges. In addition to improving LLM performance through ICL, we demonstrate that finetuning small-scale LLMs on our constructed high-quality instruction dataset leads to significant improvements of up to 6.53\% over the leading close-source LLM GPT-4. These findings highlight the potential for further advancements in enhancing LLMs' hierarchical reasoning capabilities.

%% file: 9-appendix.tex
\newpage
\section{Author Contributions}
    The authors’ contributions are: \textbf{Zhuohang Jiang}: benchmark design, the coding leader, and manuscript writing. \textbf{Pangjing Wu:} implements Binary Tree scenario tasks, evaluation, and manuscript writing. \textbf{Ziran Liang:} implements Multiple Tree scenario tasks and manuscript writing. \textbf{Peter Q. Chen:} implements JSON scenario tasks and manuscript writing. \textbf{Xu Yuan:} implements Paper scenario tasks and manuscript writing. \textbf{Ye Jia:} implements Code scenario tasks and manuscript writing. \textbf{Jiancheng Tu:} implements Formula scenario tasks and manuscript writing. \textbf{Chen Li:} project advising. \textbf{Peter H.F. Ng:} project advising.
    \textbf{Qing Li:} General project supervisor and manuscript writing.

\section{Modular Codebase}
\label{app:codebase}

\begin{figure*}[t!]
    \centering
    \includegraphics[width=0.75\linewidth]{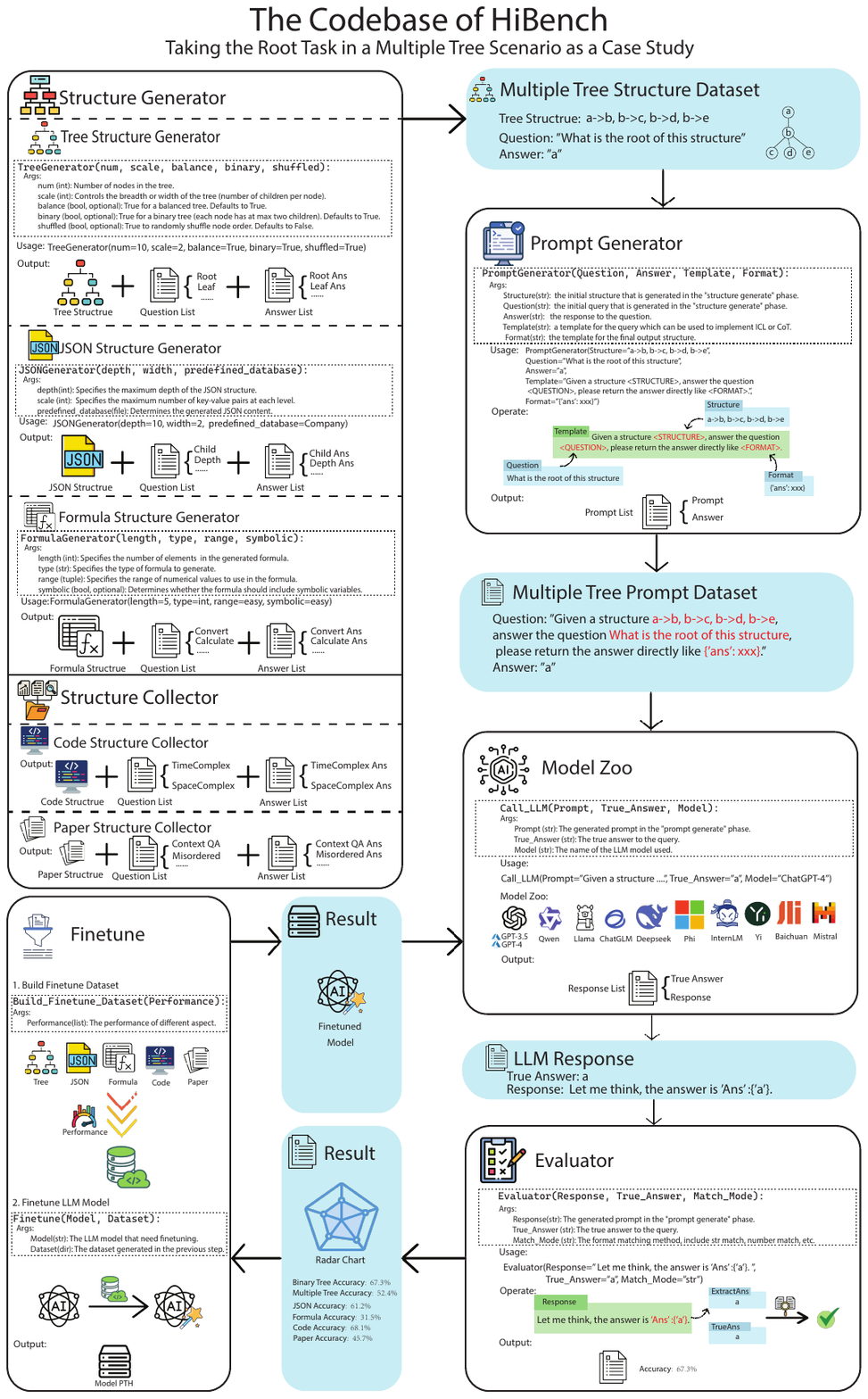}
    \caption{The Modular Codebase of HiBench.}
    \label{fig:Codebase}
\end{figure*}

Our benchmark code is designed in a modular fashion, which facilitates the swift construction of new benchmarks. As Fig~\ref{fig:Codebase} shown, the codebase is divided into five distinct modules: the \textbf{Hierarchical Structure Construction},\textbf{Query Generator}, \textbf{Prompting}, \textbf{Model Zoo}, \textbf{Evaluator}, and \textbf{Finetune}.

\subsection{Hierarchical Structure Construction}
We first employ a hierarchical structure constructor to generate a hierarchical structure. Precisely, in Binary Tree and Multiple Tree scenarios, the complexity of hierarchical structures is controlled by the the number of leaf nodes, \emph{a.k.a} width, and tree hight~\emph{a.k.a} depth.

\subsection{Query Generator}
\label{sec:}
\begin{figure}[h!]
    \centering
    \includegraphics[scale=0.42]{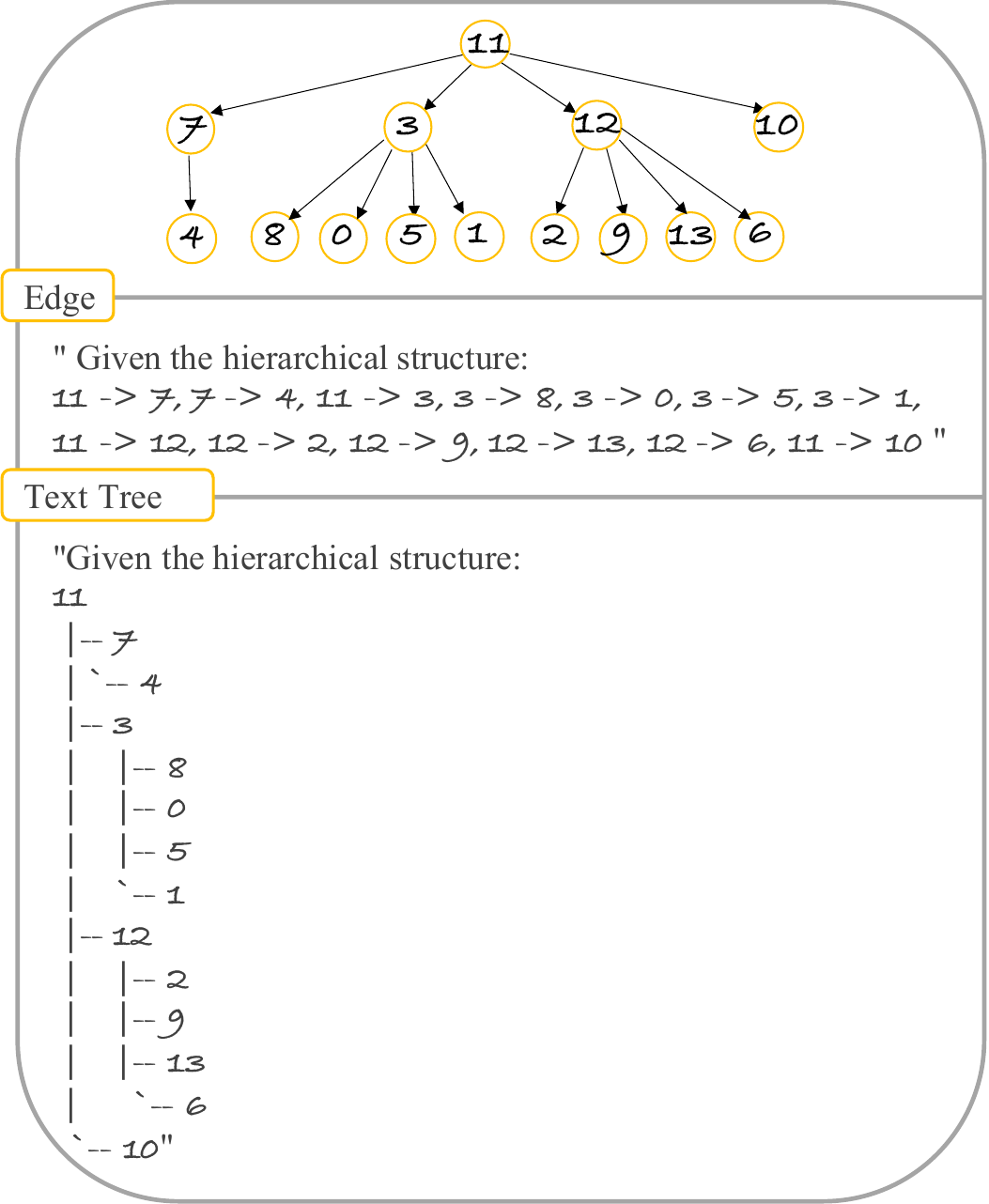}
    \caption{Different Representations of Hierarchical Tree Structure in Fundamental Task.}
    \label{fig:Task1_repre}
\end{figure}

Dataset Generator is a tool for automated dataset generation, including \textit{Binary Tree QA Generator}, \textit{Multiple Tree QA Generator}, \textit{JSON QA Generator} and \textit{Formula QA Generator}, which can flexibly generate data of different difficulty levels according to the parameters set by the user. The level of task data is determined according to the parameters set by the user. For example, we tested LLMs' understanding of various representations using tree representations, including edge and hierarchical representations. An Example has been shown in Figure~\ref{fig:Task1_repre}.

\subsection{Prompting}
Dataloader is an efficient tool dedicated to dataset organization and query generation, which mainly consists of two parts: \textit{Data Management} and \textit{Query Generator}.
\begin{itemize}
    \item \textbf{Data Management} We provide configuration parameters to manage our dataset, and subsequent evaluations are performed based on these configuration files. 

    \item \textbf{Query Generator} To reduce system coupling, we split the query and QA modules to achieve a modularized design, allowing quick modifications and adjustments. Specifically, we unified the dataset format and generated an LLM-matched alpaca prompt format for queries driven by configuration files. This design meets the needs of many subsequent tasks and has strong scalability and flexibility to respond to future iterations and optimize functionality effectively.
\end{itemize}

\subsection{Model Zoo}
Our Model Zoo is extensively integrated with resources from Hugging Face, Azure, and Fireworks, covering 20 LLMs from 10 families. We have standardized all models' input and output interfaces through a unified design, allowing for flexible invocation of different models through parameter configuration only. The data imported by the Dataloader can be directly input into the LLM model, and the model's response can be quickly obtained, thus allowing efficient and convenient model application and extension.
In the Hibench benchmarking process, two proprietary models, GPT-3.5 and GPT-4, were selected for evaluation. Additionally, 18 open-source models of various sizes were chosen from nine different families. These open-source models include Llama, Qwen, DeepSeek, ChatGLM, InternLM, BaiChuan, Yi, Phi, and Mistral.

\subsection{Evaluator}
Our evaluator mainly consists of an answer extractor and an answer evaluator. Specifically, we designed five types of matching patterns to fit different types of outputs accurately. By comparing the answers generated by LLM with the standard answers, we calculate the accuracy and hit rate and categorize the evaluation results in five dimensions. This all-encompassing evaluation approach can comprehensively measure the performance of LLM in different tasks, which provides strong support for in-depth profiling of model performance.

\subsection{Finetune}
Our fine-tuning module, built upon the Unsloth framework, has been developed to optimize the performance of LLMs through targeted fine-tuning. This approach aims to enhance the LLMs' capabilities in hierarchical comprehension, thereby improving their capability to effectively process and understand complex, layered information structures.

\section{Benchmark Detail}

\label{app:dataset}

\begin{figure*}[h!]
    \centering
    \includegraphics[scale=0.44]{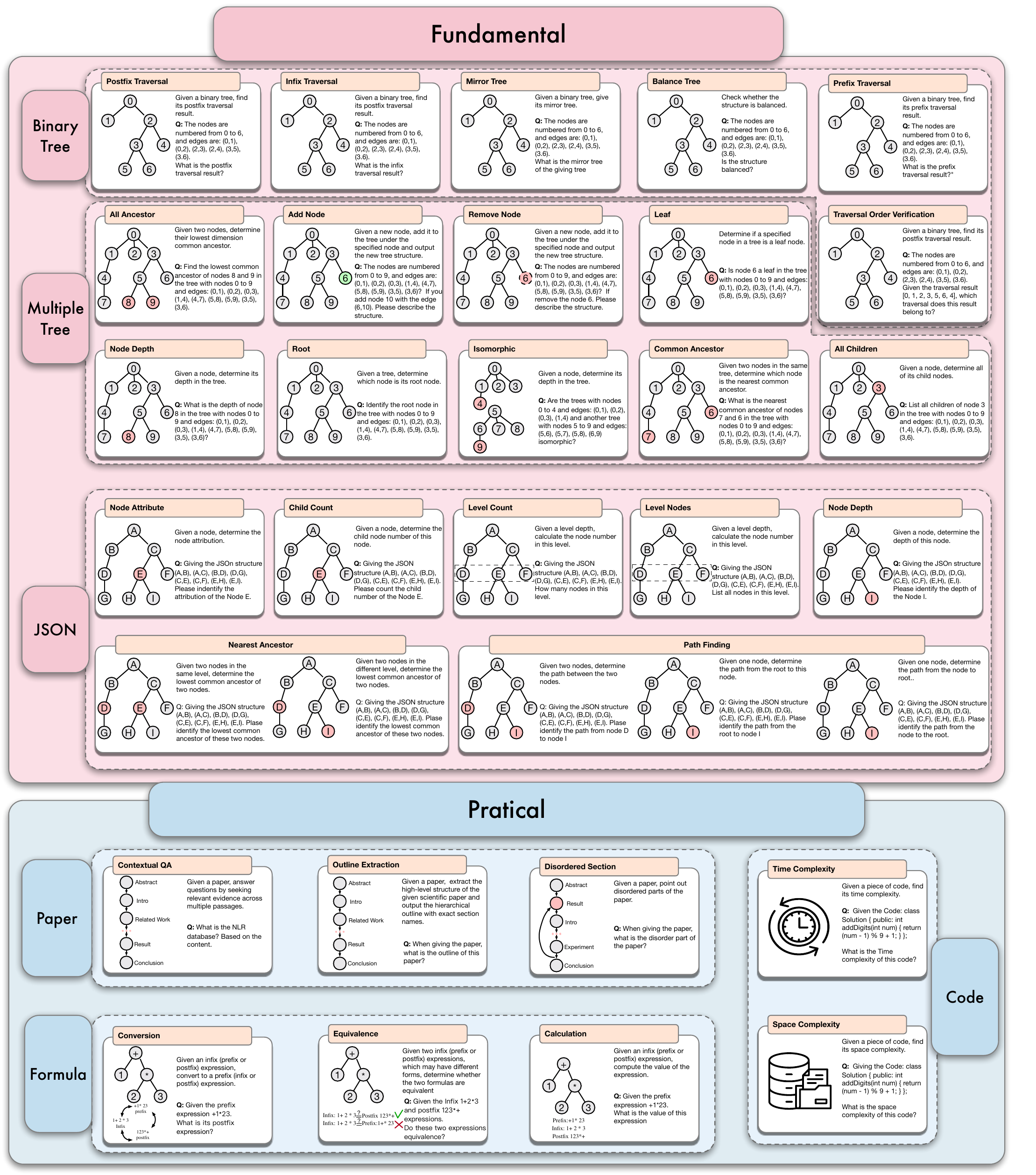}
    \caption{HiBench Tasks: A Comprehensive Introduction.}
    \label{fig:TaskFIgure}
\end{figure*}

In this section, we will define the scope of the task in depth and give a detailed description of the dataset used. We will describe our task comprehensively and in-depth in two dimensions: the \textit{Fundamental Aspect} and the \textit{Practical Aspect}. Detailed information is presented in Fig.~\ref{fig:TaskFIgure}.
\subsection{Fundamental Aspect}
In exploring the fundamental aspect of the task, we focus on evaluating the capability of LLMs to understand different underlying hierarchical structures, such as \textit{Binary Tree Scenario}, \textit{Multiple Tree Scenario}, and \textit{JSON Scenario}.
\subsubsection{Binary Tree Scenario}
Binary Tree, a widely used basic hierarchical structure, has up to two children per node. Based on this feature, we can propose a series of challenging binary tree scenario tasks designed to deeply test LLM's capability to understand and manipulate this simple but critical data structure.

\paragraph{Task Definition}
The Binary Tree Scenario covers six tasks that focus on evaluating the capabilities of LLM models in three key areas: structural modification and relationship understanding. 
\begin{itemize}
    \item \textit{Mirror Tree}: Given a binary tree, give its mirror tree.
    \item \textit{Balance}: Given a binary tree, Determine if it is balanced.
    \item \textit{Prefix Traversal}: Given a binary tree, find its prefix traversal result
    \item \textit{Postfix Traversal}: Given a binary tree, find its postfix traversal result
    \item \textit{Infix Traversal}: Given a binary tree, find its infix traversal result
    \item \textit{Traversal Order Verification}: Given a binary tree and a specific traversal result, it is necessary to determine which traversal the result belongs to.
\end{itemize}
\paragraph{Random Binary Tree Generator}
We have designed a random binary tree generator to comprehensively evaluate the LLM model's understanding capabilities over multiple tree data. We can generate multiple tree structures with different difficulty levels by defining the depth $D$ and the number $N$ of nodes.
\paragraph{Binary Tree Dataset Statistics}
\begin{table}[h!]
\caption{Structure Statistics of Binary Tree Scenario.}
\label{tab:basic_task_dataset}
\begin{tabular}{ccccc}
    \toprule
    Complexity& \#Structure & \#Node & \#Layer & \#Degree \\
    \midrule
    {Easy} & 48 & {2\textasciitilde 15}& {2\textasciitilde 4}& 2\\
    {Medium} & 54 & {16\textasciitilde 255}& {5\textasciitilde 7}& 2\\
    {Hard} & 36 & {256\textasciitilde 511}& {8\textasciitilde 9}& 2\\
    \bottomrule
\end{tabular}
\end{table}

\subsubsection{Multiple Tree Scenario}
As a hierarchical structure with high complexity, a multiple tree can accurately measure an LLM's basic reasoning of hierarchical architectures. With this in mind, we have crafted a series of challenging tasks designed to comprehensively test the LLM's capability to understand and manipulate hierarchical data.
\paragraph{Task Definition}
 The Multiple Tree Scenario covers nine tasks that focus on evaluating the capabilities of LLM models in three key areas: structural modification, relationship understanding, and structural analysis. 
\begin{itemize}
    \item \textit{Add Node}: Given a new node, add it to the tree under the specified node and output the new tree structure.
    \item \textit{Remove Node}: Delete the specified node in the tree and output the new tree structure. If the deleted node is the root node, the first child of the root node needs to be taken as the new root node, and other children are taken as children of the new root node.
    \item \textit{Leaf}: Determine if a specified node in a tree is a leaf node.
    \item \textit{All Ancestor}: Given two nodes, determine their lowest dimension common ancestor.
    \item \textit{All Children}: Given a node, determine all its child nodes.
    \item \textit{Common Ancestor}: Given two nodes in the same tree, determine which node is the nearest common ancestor.
    \item \textit{Node Depth}: Given a node, determine its depth in the tree.
    \item \textit{Root}:  Given a tree, determine which node is its root node.
    \item \textit{Isomorphic}: Given two trees, determine if they are isomorphic,~\emph{i.e.}, structurally identical.
\end{itemize}
\paragraph{Random Multiple Tree Generator}
We have designed a random multiple tree generator to comprehensively evaluate the LLM model's understanding capabilities over multiple tree data. We can generate numerous tree structures with different difficulty levels by defining the maximum out-degree $W$ of the tree, the depth $D$, and the number $N$ of nodes.

\paragraph{Multiple Tree Dataset Statistics}
The HiBench designs three datasets of different sizes that vary in nodes, node degree, and depth.
As shown in Table~\ref{tab:basic_task_dataset_multi}, the number associated with each dataset of different difficulty levels is the size of the dataset, the number of nodes in the tree, the number of layers in the tree, and the out-degree of the nodes in the tree, respectively.

\begin{table}[h!]
\caption{Structure Statistics of Multiple Tree Scenario.}
\label{tab:basic_task_dataset_multi}
\begin{tabular}{ccccc}
    \toprule
    Complexity& \#Structure & \#Node & \#Layer & \#Degree \\
    \midrule
    {Easy} & 33 & {2\textasciitilde 13}& {2\textasciitilde 3}& {2\textasciitilde 3}\\
    {Medium-1} & 36 & {2\textasciitilde 13}& {2\textasciitilde 3}& {3\textasciitilde 4}\\
    {Medium-2} & 36 & {3\textasciitilde 34}& {3\textasciitilde 4}& {2\textasciitilde 3}\\
    {Hard-1} & 36 & {4\textasciitilde 32}& {2\textasciitilde 3}& {5\textasciitilde 6}\\
    {Hard-2} & 36 & {13\textasciitilde 212}& {5\textasciitilde 6}& {2\textasciitilde 3}\\
    \bottomrule
\end{tabular}
\end{table}

Additionally, we tested LLMs' understanding of various representations using tree representations, including edge and hierarchical representations. 
An Example has been shown in Figure~\ref{fig:Task1_repre}.

\subsubsection{JSON Scenario}
JSON (JavaScript Object Notation) is a lightweight semi-structured data exchange format. It is text-based and adopts a structure consisting of key/value pairs and ordered lists, which enables the effective representation of hierarchical information. Given this, we designed a series of challenging JSON scenario tasks to test the capability of LLM to reason about hierarchical data.
\paragraph{Task Definition}
 The JSON Scenario covers seven tasks that focus on evaluating the capabilities of LLM models in two key areas: relationship understanding and structural analysis. 
\begin{itemize}
    \item \textit{Path Finding}:
    \begin{itemize}
        \item \textit{Path Between Nodes}: Given two nodes, determine the path between the two nodes.
        \item \textit{Path Down to Up}: Given a node, determine the path from the root to the node.
        \item \textit{Path Up to Down}: Given a node, determine the path from the node to the root.
    \end{itemize}
    \item \textit{Nearest Ancestor}:
    \begin{itemize}
        \item \textit{Shared Ancestor Same Level}: Given two nodes in the same level, determine the lowest common ancestor of two nodes.
        \item \textit{Shared Ancestor Diff Level}: Given two nodes in different levels, determine the lowest common ancestor of two nodes.
    \end{itemize}
    \item \textit{Node Depth}: Given a node, determine the depth of this node.
    \item \textit{Child Count}: Given a node, determine the child node number of this node.
    \item \textit{Level Count}: Given a level depth, calculate the node number in this level.
    \item \textit{Level Nodes}: Given a level depth, determine all nodes in this level.
    \item \textit{Node Attribute}: Given a node, determine the node attribution.
\end{itemize}
\paragraph{Random JSON Generator}
We designed a random JSON generator to comprehensively evaluate the LLM model's capability to understand multi-tree data. We can generate multiple JSON structures with different difficulty levels and meanings by defining the maximum degree $W$, depth $D$, and node meaning $M$ of JSON.
\paragraph{JSON Dataset Statistics} The JSON dataset comprises seven types of questions derived from two categories of datasets: normal and nonsense. The normal dataset contains JSON structures with realistic and meaningful semantics, such as university names that closely resemble real-life entities. In contrast, the nonsense dataset includes JSON structures with meaningless content, such as departments named with random words or numbers. Please refer to Table~\ref{tab:json_dataset} for detailed information about the JSON dataset. 

\begin{table}[h!]
\caption{Statistics of JSON Datasets.}
\label{tab:json_dataset}
\begin{tabular}{ccc}
    \toprule
    Complexity & \#Depth & \#Width \\
    \midrule
    {Small} & {4}& {2}\\
    {Medium-1}& {5}& {2}\\
    {Medium-2} & {4}& {4}\\
    {Large-1} & {6}& {2}\\
    {Large-2} & {4}& {6}\\
    \bottomrule
\end{tabular}
\end{table}

\subsection{Practical Aspect}
In exploring the application level of hierarchical understanding, we focus on evaluating the capability of LLM to understand hierarchical structures in different application scenarios, such as \textit{Code Scenario}, \textit{Formula Scenario} and \textit{Paper Scenario}.
\subsubsection{Code Scenario}
Code is the basic unit of composition of a computer program with a strict hierarchical structure, where blocks of code at different levels clearly define the hierarchical relationships to which they belong, utilizing indentation, curly brackets, etc. Based on this, we propose a series of challenging Code Scenario tasks designed to deeply test LLM's capability to understand and manipulate this complex hierarchical structure. 
\paragraph{Task Definition}
 The Code Scenario covers two tasks that focus on evaluating the capabilities of LLM models in Analytical Reasoning areas.
\begin{itemize}
    \item \textit{Space Complexity}: Given a piece of code, find its space complexity.
    \item \textit{Time Complexity}: Given a piece of code, find its time complexity.
\end{itemize}
\paragraph{Code Dataset Collection}
The dataset comprises Python
and C++ codes sourced from GitHub repositories,
specifically targeting LeetCode problems. The questions on LeetCode cover a broad spectrum of topics, ranging from basic data structures to complex algorithms, making them highly representative of real-world programming challenges.
\paragraph{Code Dataset Statistics}
The dataset consists of 100 Python and 100 C++ scripts, each containing two questions about time and space complexity. Please refer to Table \ref{tab:code_datasets} for the details of the datasets.

\begin{table}[]
\caption{Statistics of Code Understanding Datasets.}
\label{tab:code_datasets}
\begin{tabular}{lll}
\toprule
Programming Language   & Python      & C++    \\ 
\midrule
Script Content         & LeetCode & LeetCode \\
Script Length          & 11 - 86 lines   & 6 - 37 lines       \\
No. of scripts         & 100       & 100    \\ 
\midrule
Question no. of Type 1 & 100       & 100      \\
Question no. of Type 2 & 100       & 100      \\ 
\bottomrule
\end{tabular}
\end{table}

\subsubsection{Formula Scenario}
Formulas are an essential tool for expressing complex relationships and laws in scientific research and engineering applications. They consist of elements such as symbols, variables, and operators, forming a form of expression with a strict hierarchical structure. Based on this, we propose a series of challenging formula scenario tasks designed to test the capability of LLM to understand and manipulate this complex hierarchical structure deeply. 
\paragraph{Task Definition}
 The Formula Scenario covers three tasks that focus on evaluating the capabilities of LLM models in Analytical Reasoning areas.
\begin{itemize}
    \item \textit{Calculation}: Given an infix (prefix or postfix) expression, compute the value of the expression.
    \item \textit{Conversion}: Given an infix (prefix or postfix) expression, convert it to a prefix (infix or postfix) expression.
    \item \textit{Equivalence}: Given two infix (prefix or postfix) expressions, which may have different forms, determine whether the two formulas are equivalent.
\end{itemize}
\paragraph{Random Formula Generator}
We design a stochastic Formula generator to comprehensively evaluate the LLM model's capability to comprehend multi-tree data. Multiple formula structures with different difficulty levels can be generated by defining the length $L$, symbolic complexity $S$, and numeric complexity $N$ of the formulas.
\paragraph{Formula Dataset Statistics}
As shown in Table~\ref{Tab:FormulaDiff}, we constructed a system of formulas of different complexity in the Formula task in four dimensions: Type, Range, Symbolic, and Length, and generated the corresponding prefix, infix, and postfix expressions. These expressions are systematically assigned to Conversion, Calculation, and Equivalence tasks. Specifically, in the Formula scenario, the Conversion task contains 5022 queries, the Calculation task contains 2511 queries, and the Equivalence task contains 7533 queries, resulting in a total of 15066 task instances.

\begin{table}[]
\caption{Multidimensional Formula Complexity System.}
\label{Tab:FormulaDiff}
\begin{tabular}{cccc}
\toprule
Complexity & Easy   & Medium   & Hard                     \\
\midrule
Type       & int    & -        & float                    \\
Range      & 1 - 10 & -50 - 50 & -100 - 100               \\
Symbolic   & + - * /   & + - * / ( )   & + - * / ( ) \textasciicircum{} \\
Length     & 3      & 6        & 9          \\             
\bottomrule
\end{tabular}
\end{table}

\subsubsection{Paper Scenario}
A dissertation typically consists of multiple levels, including an introduction, methods, results, discussion, and conclusion sections, with a strong hierarchical relationship between these sections. For this reason, we propose a series of challenging paper scenario tasks designed to deeply test LLM's capability to understand and manipulate this complex hierarchical structure.
\paragraph{Task Definition}
 The Paper Scenario covers three tasks that focus on evaluating the capabilities of LLM models in Textual Reasoning areas.
\begin{itemize}
    \item \textit{Contextual QA}: Asking LLMs to answer questions by seeking relevant evidence across multiple passages.
    \item \textit{Disordered Section}: Asking LLMs to point out disordered parts of the paper.
    \item \textit{Outline Extraction}: LLMs are required to extract the high-level structure of the given scientific paper and output the hierarchical outline with exact section names.
\end{itemize}
\paragraph{Paper Dataset Collection}
We chose the Qasper dataset \cite{dasigi2021dataset} as the source of scientific papers to construct our task's training and test sets.
Specifically, Qasper covers 1,585 papers and 5,049 questions, which are asked by natural language processing practitioners who have only read the title and abstract of the corresponding paper and aim to obtain information from the full text of the paper. 
After filtering those questions requiring information on figures and tables, we get 3,570 question-answer pairs for our contextual question answering. 
We convert these papers into document trees by analyzing their structures before feeding them to the LLMs. 

\paragraph{Paper Dataset Statistics}

\begin{table}[]
\caption{Statistics of Paper Understanding Datasets.}
\label{tab:paper-understanding-dataset}
\begin{tabular}{cccc}
\toprule
Task & Splits & \#Samples & \#Avg Tokens \\
\midrule
\multirow{3}{*}{\makecell[c]{Outline \\ Extraction}} & train & 179 & 6045 \\
 & dev & 42 & 5865 \\
 & test & 80 & 5528 \\
 \midrule
\multirow{3}{*}{\makecell[c]{Disordered \\ Section}} & train & 354 & 5981 \\
 & dev & 85 & 5815 \\
 & test & 163 & 5569 \\
 \midrule
\multirow{3}{*}{\makecell[c]{Contextual \\ QA}} & train & 404 & 5975 \\
 & dev & 144 & 5694 \\
 & test & 237 & 5336 \\
\bottomrule
\end{tabular}
\end{table}

As Table~\ref{tab:paper-understanding-dataset} shows, this dataset focuses on tasks related to academic paper understanding, encompassing three sub-tasks: Outline Extraction, Disordered Section Identification, and Contextual Question Answering. It aims to evaluate the capability of models to comprehend the structure and content of scholarly articles. The dataset is divided into training, development, and testing sets, with each sub-task dataset containing thousands of samples and an average token count ranging from 5,300 to 6,000.

\section{Experimental Setup}
\label{app:setup}
\subsection{Model Introduction}
\input{tabels/companyinfo}

In the HiBench, we evaluated 20 LLMs from 10 model families. These models are categorized into four main groups, namely the \textit{GPT family}, the \textit{Llama family}, the \textit{Qwen family}, and \textit{Other Open-source Models}. Table~\ref{tab:company info} shows the information of different models.
\begin{itemize}
    \item GPT Family: The GPT family is a series of advanced large-scale language models developed by OpenAI and known for their superior natural language processing capabilities.GPT-3.5 and GPT-4 perform well in several benchmarks. In our HiBench benchmarks, we have chosen GPT-3.5 and GPT-4 for evaluation to fully demonstrate the performance of the GPT family in different tasks.
    \item Llama Family: The Llama Family is a family of open-source large-scale language models from Meta, noted for their excellent multilingual support, extended context processing capabilities, and efficient architectural design. We selected five models for our benchmarking: Llama-3.2-1B, Llama-3.2-3B, Llama-3.1-8B, Llama-3.1-70B, and Llama-3.1-405B. 
    \item Qwen Family: Qwen Family is a family of open-source large-scale language models developed by Alibaba Dharma Institute, which is highly regarded for its strong multilingual support, extended context processing capabilities, and efficient architectural design. For our benchmarks, we have selected the following models from the Qwen2.5 family: Qwen2.5-0.5B, Qwen2.5-1.5B, Qwen2.5-3B, Qwen2.5-7B, QwQ-32B, and Qwen2.5-72B.
    \item Other Open-source LLMs: Our benchmark also covers representative models from seven other model families, including ChatGLM, Phi, InternLM, Yi, Baichuan, Mistral, and DeepSeek, which demonstrate their respective strengths in multilingual comprehension, complex reasoning, and code generation tasks. By introducing these diverse models, our benchmarks provide a more comprehensive picture of current large-scale language models' performance levels and trends.
\end{itemize}

\subsection{Hyperparameters}
In this model setup, we make specific configurations for key parameters to ensure the model runs as intended. Expressly, we set the temperature to 0 to ensure the stability and consistency of the generated content. At the same time, we fix the seed of the local model to 0 to facilitate the reproduction and validation of the results. In addition, we set the new token length to 2048, which allows the model to generate longer textual content to support complex tasks such as writing lengthy articles, generating detailed reports, or engaging in in-depth dialogues.

\subsection{Hardware Configurations} 
Experiments involving open-source LLMs smaller than 14B parameters were completed on multiple independent workstations, each equipped with two NVIDIA GeForce RTX 3090 or RTX A6000 GPUs. In contrast, experiments with LLMs larger than 14B parameters were conducted by calling API from Fireworks platform\footnote{https://fireworks.ai} and Azure platform\footnote{https://azure.microsoft.com}.

\subsection{Evaluation Details}
Algorithm ~\ref{Code:match} shows an example of the evaluation process. All experiments are based on the codebase we developed. We first generate a complete query request (query) using a file and a query template, feed it into a large language model (LLM), and wait for the model's response. Subsequently, we format and extract the response data from the LLM, compare it with the correct answer, and finalize the evaluation.
\input{tabels/matchcode}

\begin{figure*}[h!]
    \centering
    \includegraphics[scale=0.5]{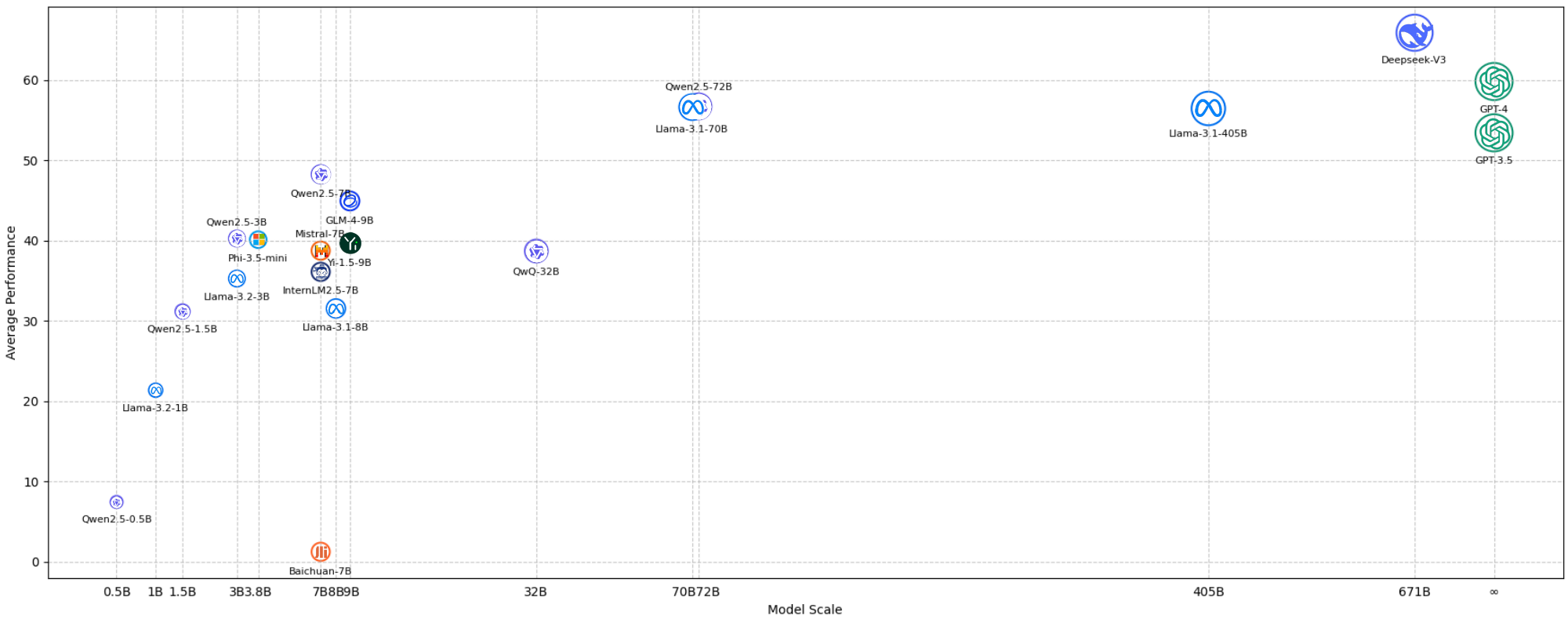}
    \caption{Different LLM's Capability of Hierarchical Structure Reasoning in HiBench.}
    \label{fig:scalinglaw}
\end{figure*}

\section{Result Analysis}
This section systematically analyses the performance of Large Language Models (LLMs) in HiBench benchmarks, focusing on their performance in several key areas.
\label{app:result}
\subsection{Main Results}
Figure~\ref{fig:scalinglaw} illustrates the relationship between hierarchical reasoning ability and model size for different Large Language Models (LLMs) in HiBench tests. It can be observed that there is a positive correlation between model size and average performance, i.e., as the number of model parameters increases, the model's performance on hierarchical reasoning tasks tends to improve. For example, models with billions or even more parameters, such as Llama-3.1-70B and Qwen2.5-72B, show higher average performance scores, while models with a smaller number of parameters perform relatively lower.

In addition, the figure reveals significant differences in performance between models. Some large-scale models, such as DeepSeek-V3, GPT-4, and Phi-3.5-mini, are not only large in scale but also show excellent performance in inference tasks. This suggests that for application scenarios requiring complex reasoning capabilities, choosing these large-scale, high-performance models may be more appropriate. However, the figure also shows that some medium-sized models have the potential to improve their performance, suggesting that in addition to size, the model's structural design and training strategy are also key factors affecting its reasoning ability.

\subsection{Finetune Result}

\begin{table}[]
\caption{Instruction Finetuning Performance of Qwen2.5-7B and Llama-3.1-8B on HiBench.}
\label{Tab:finetune2}
\begin{tabular}{ccccc}
\toprule
\multicolumn{1}{c}{\multirow{2}{*}{Scenario}} & \multicolumn{2}{c}{Qwen2.5-7B} & \multicolumn{2}{c}{Llama-3.1-8B} \\ \cmidrule(lr){2-3} \cmidrule(lr){4-5} 
\multicolumn{1}{l}{}                      & Origin        & Finetune       & Origin         & Finetune        \\ 
\midrule
Binary Tree                               & 33.11         & 67.34          & 14.74          & 66.70           \\ 
Multiple Tree                             & 54.86         & 83.20          & 28.76          & 82.61           \\
JSON                                      & 39.00         & 46.67          & 34.18          & 41.45           \\ 
Formula                                   & 69.25         & 80.05          & 61.75          & 78.64           \\ 
Code                                      & 51.17         & 84.75          & 33.63          & 82.50           \\ 
Paper                                     & 42.08         & 18.34          & 16.23          & 5.65            \\ 
\midrule
Ave.                                      & 48.25         & \textbf{63.39}          & 31.55          & \textbf{59.58}           \\ 
\bottomrule
\end{tabular}
\end{table}

As Table~\ref{Tab:finetune2} shows, instruction fine-tuning significantly enhanced the performance of Qwen2.5-7B and Llama-3.1-8B on the HiBench benchmark across tasks such as Binary Tree, Multiple Tree, JSON, Formula, and Code, with average scores improving from 48.25 to 63.39 for Qwen2.5-7B and from 31.55 to 59.58 for Llama-3.1-8B. However, in the Paper task, both models exhibited a notable decline in performance, with Qwen2.5-7B dropping from 42.08 to 18.34 and Llama-3.1-8B from 16.23 to 5.65. This decline is attributed to the Paper task's emphasis on textual comprehension rather than merely hierarchical information processing, rendering fine-tuning focused on hierarchical capabilities less effective. Overall, Qwen2.5-7B consistently outperformed Llama-3.1-8B across all tasks, while the latter demonstrated greater relative improvement potential in tasks with lower initial performance.

\subsection{Binary Tree}
\input{tabels/binarytree}
Table~\ref{Tab:binarytreeLLMPerformance} illustrates the leaderboard for the HiBench binary tree scenario, showing the performance of the Model name for different Model Families on multiple tasks. As can be seen from the table, DeepSeek's DeepSeek-V3 model performs best on all tasks, ranking first with an average score of 70.23. It is followed by Meta's Llama-3.1-405B model with an average score of 64.49, and in third place is OpenAI's GPT-4 model with an average score of 56.29. The best performer among the open-source models is Llama-3.1-405B, while the best performer among the closed-source models is GPT-4.

\subsection{Multiple Tree}
\input{tabels/multipletree}
Table~\ref{Tab:multipletreeLLMPerformance} illustrates the leaderboard for HiBench multi-tree scenarios, showing the performance of the Model name on multiple tasks for different Model Families. The table includes both Closed-Sourced and Open-Sourced models. The performance of each model is given in the form of scores on different tasks. In addition, the table provides the Average and overall Rank for each model. As can be seen from the table, Qwen's Qwen2.5-32B model performed the best among all the tasks, ranking first with an average score of 74.12, followed by OpenAI's GPT-4 model with an average score of 73.64 and third place was DeepSeek's DeepSeek-V3 model with an average score of 72.59.

\subsection{JSON}
\input{tabels/json}
Table~\ref{Tab:JSONLLMPerformance} illustrates the leaderboard for HiBench JSON scenarios, showing the performance of different Model Families and Model names on multiple tasks. As you can see from the table, DeepSeek's DeepSeek-V3 model performs the best on all tasks, ranking first with an average score of 77.65. This is followed by Qwen's Qwen2.5-32B model with an average score of 73.80, and in third place is Qwen's Qwen2.5-72B model with an average score of 70.50. The best performer among the open-source models is Qwen2.5-32B, while the best performer among the closed-source models is GPT-4. The table also shows the specific scores of the different models on the various tasks, which helps to understand the strengths and weaknesses of each model in a particular task. The overall average performance score of 49.94 provides a benchmark for model performance.

\subsection{Formula}
\input{tabels/formula}
Table~\ref{Tab:formulaLLMPerformance} illustrates the leaderboard of HiBench formula scenarios showing the performance of the Model name on multiple tasks for different Model Families. The table includes both Closed-Sourced and Open-Sourced models. The performance of each model is given in the form of scores on different sub-tasks, including Calculation, Conversion, and Equivalence. In addition, the table provides the Task Average and overall Rank for each model. As can be seen from the table, DeepSeek's DeepSeek-V3 model performs the best among all tasks, ranking first with an average score of 57.40, followed by Meta's Llama-3.1-405B model with an average score of 53.37, and OpenAI's GPT-4 model in third place with an average score of 54.37. Of the open-source models, The best-performing open-source model is Llama-3.1-405B, while the best-performing closed-source model is GPT-4.

\subsection{Code}
\input{tabels/code}
Table~\ref{Tab:codrLLMPerformance} illustrates the leaderboard of HiBench code scenarios showing the performance of Model name on code-related tasks for different Model Families. The table includes both Closed-Sourced and Open-Sourced models. The performance of each model is given in the form of scores on different tasks including Space Complexity and Time Complexity. In addition, the table also provides the average task scores and the overall ranking for each model. As can be seen from the table, DeepSeek's DeepSeek-V3 model performs the best among all tasks, ranking first with an average score of 72.5, followed by Meta's Llama-3.1-405B-Instruct model with an average score of 70.5, and in third place is THUDM's GLM-4-9B-chat model with an average score of 66.5. The best-performing open-source model is Llama-3.1-405B-Instruct, while the best-performing closed-source model is GPT-4.

\subsection{Paper}
\input{tabels/paper}
Table~\ref{Tab:paperLLMPerformance} is a leaderboard of HiBench paper scenarios showing the performance of Model name on multiple tasks for different Model Families. The table includes both Close-Sourced and Open-Sourced models. The performance of each model is given in the form of scores on different tasks, including Contextual QA, Outline Extraction, and Disordered Section. In addition, the table also provides the average task scores and overall ranking for each model. From the table, it can be seen that Meta's Llama-3.1-70B model performs the best among all the tasks with an average score of 43.71, followed by DeepSeek's DeepSeek-V3 model with an average score of 42.99, and Qwen's Qwen2.5-7B model with an average score of 42.08 in the third place. open source models The best-performing open-source model is Llama-3.1-70B, while the best-performing closed-source model is GPT-4.

\section{Case Study}
\label{app:casestudy}
\subsection{Baichuan Cases}
\label{app:casestudy-baichuan}
\begin{figure*}[t]
    \centering
    \begin{subfigure}[b]{0.6\linewidth}
        \includegraphics[width=\linewidth]{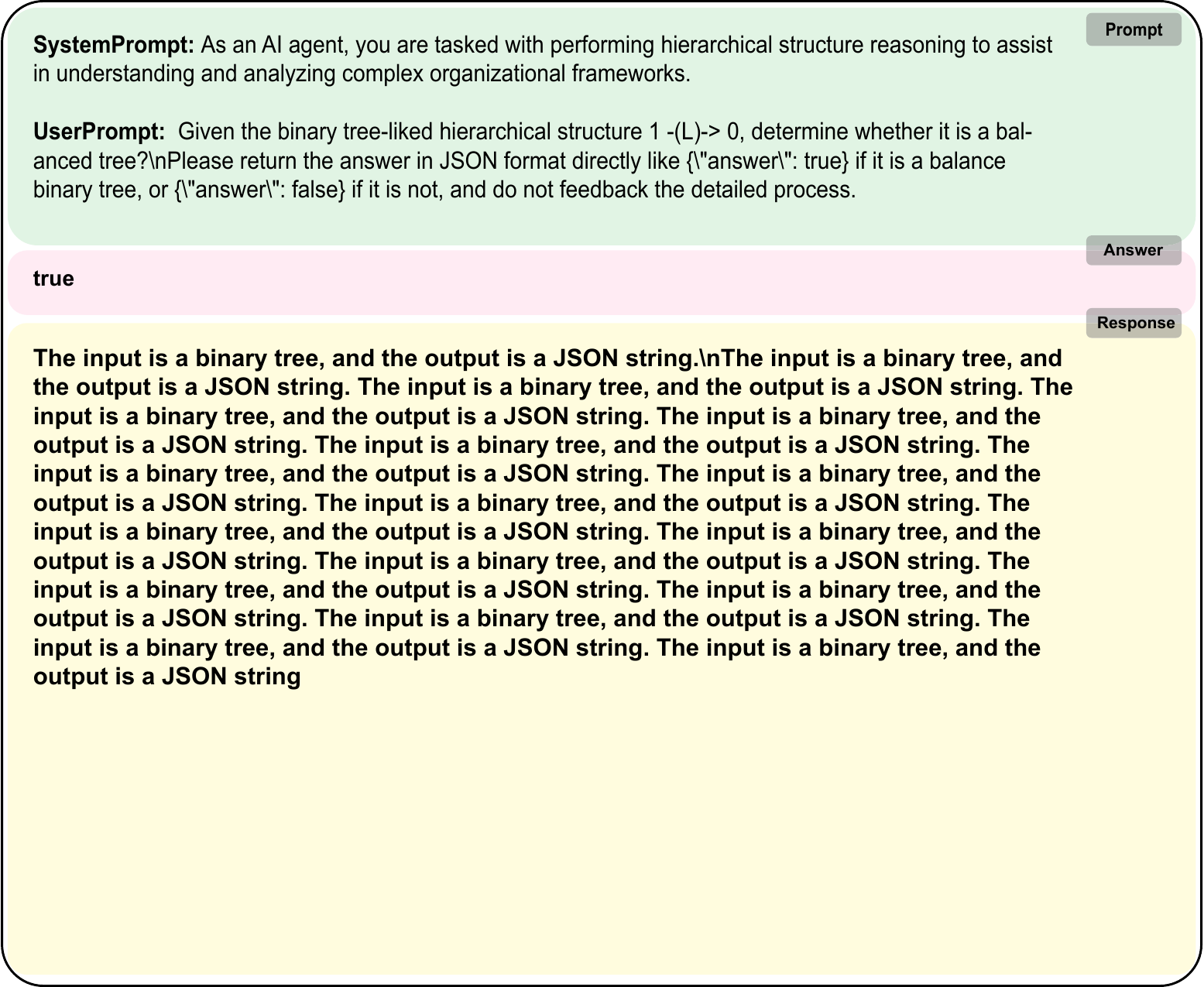}
        \caption{Case Study of Baichuan-7B on Balance Task in the Binary Tree Scenario.}
        \label{fig:case-baichuan-balance}
    \vspace{0.6cm}
    \end{subfigure}
    \begin{subfigure}[b]{0.6\linewidth}
        \includegraphics[width=\linewidth]{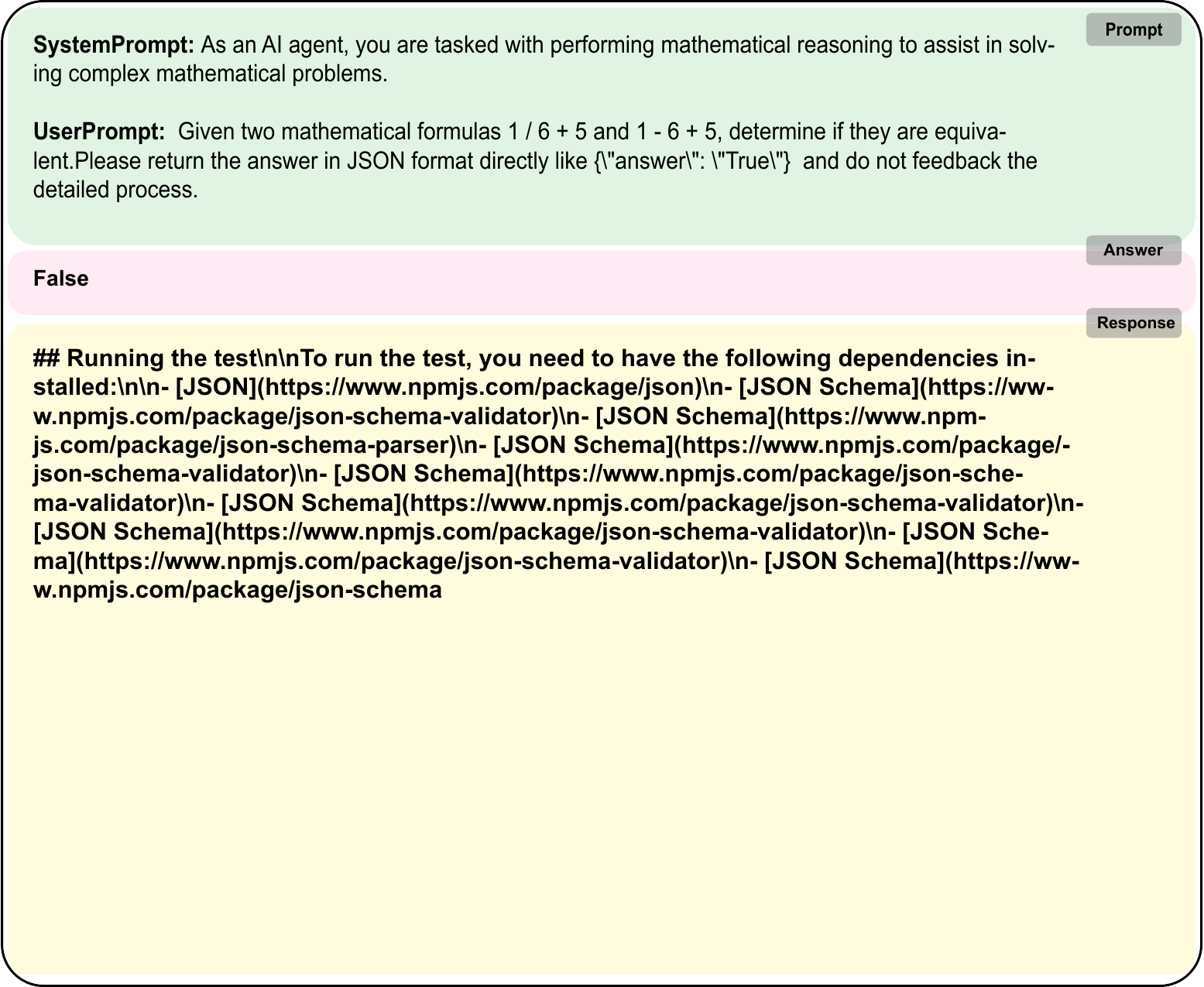}
        \caption{Case Study of Baichuan-7B on Equivalent Task in the Formula Scenario.}
        \label{fig:case-baichuan-equivalent}
    \end{subfigure}
    \caption{Case Study of Baichuan-7B on HiBench.}
    \label{fig:case-baichuan}
\end{figure*}

To explore the reasons behind Baichuan-7B's unexpectedly poor performance on HiBench, we conduct case studies on the \textit{balance} task in the Binary Tree scenario and the \textit{equivalent} task in the Formula scenario. As shown in Figure~\ref{fig:case-baichuan}, Baichuan-7B struggles to correctly understand and repeat the instructions. Further improvements are needed to enhance its ability to comprehend human instructions.

\subsection{Chain-of-Thought Cases}
\label{app:casestudy-cot}
\begin{figure*}[t]
    \centering
    \begin{subfigure}[b]{0.6\linewidth}
        \includegraphics[width=\linewidth]{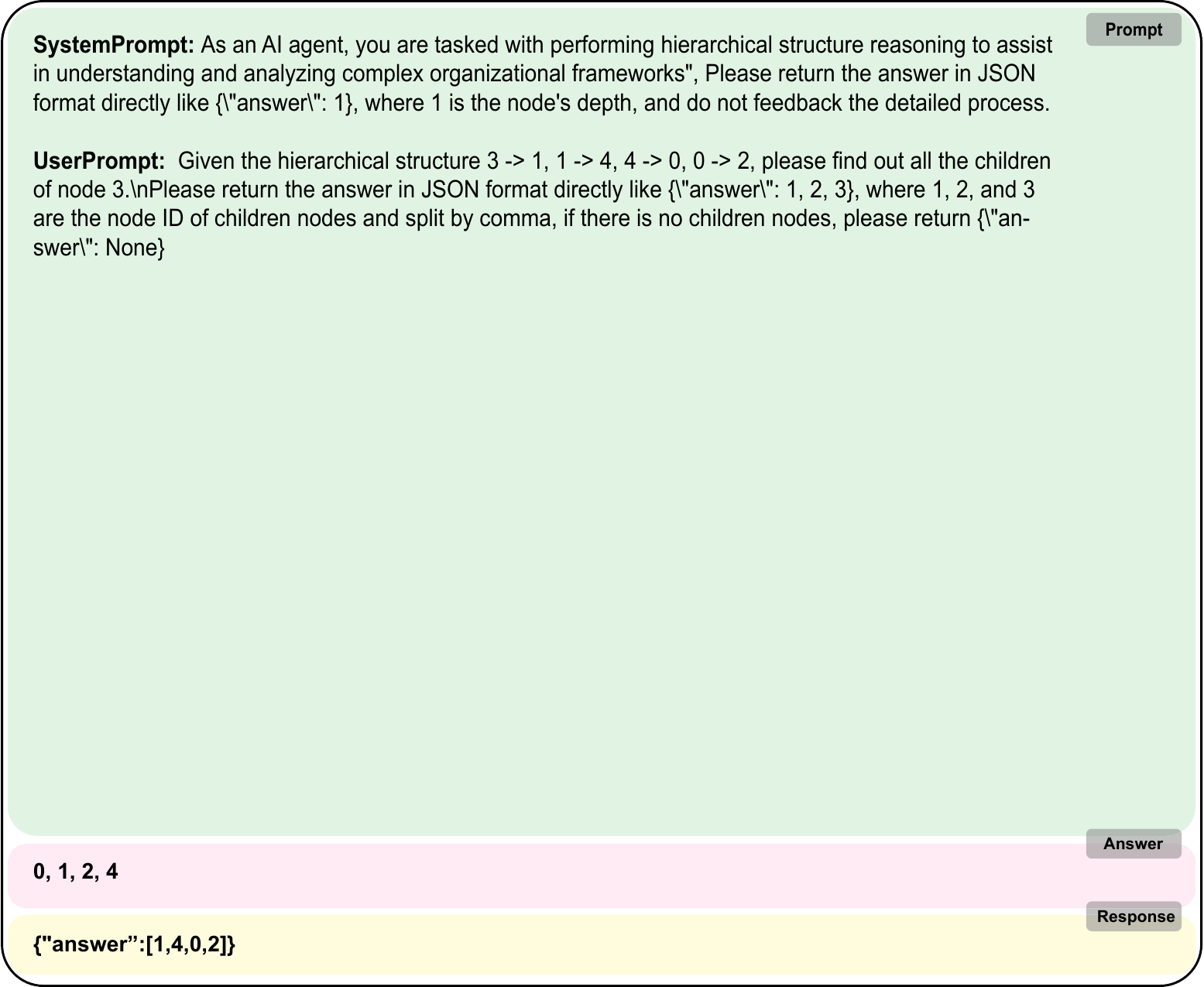}
        \caption{Non-CoT Case over All Children Task.}
        \label{fig:case-cot-non-cot}
    \vspace{0.6cm}
    \end{subfigure}
    \begin{subfigure}[b]{0.6\linewidth}
        \includegraphics[width=\linewidth]{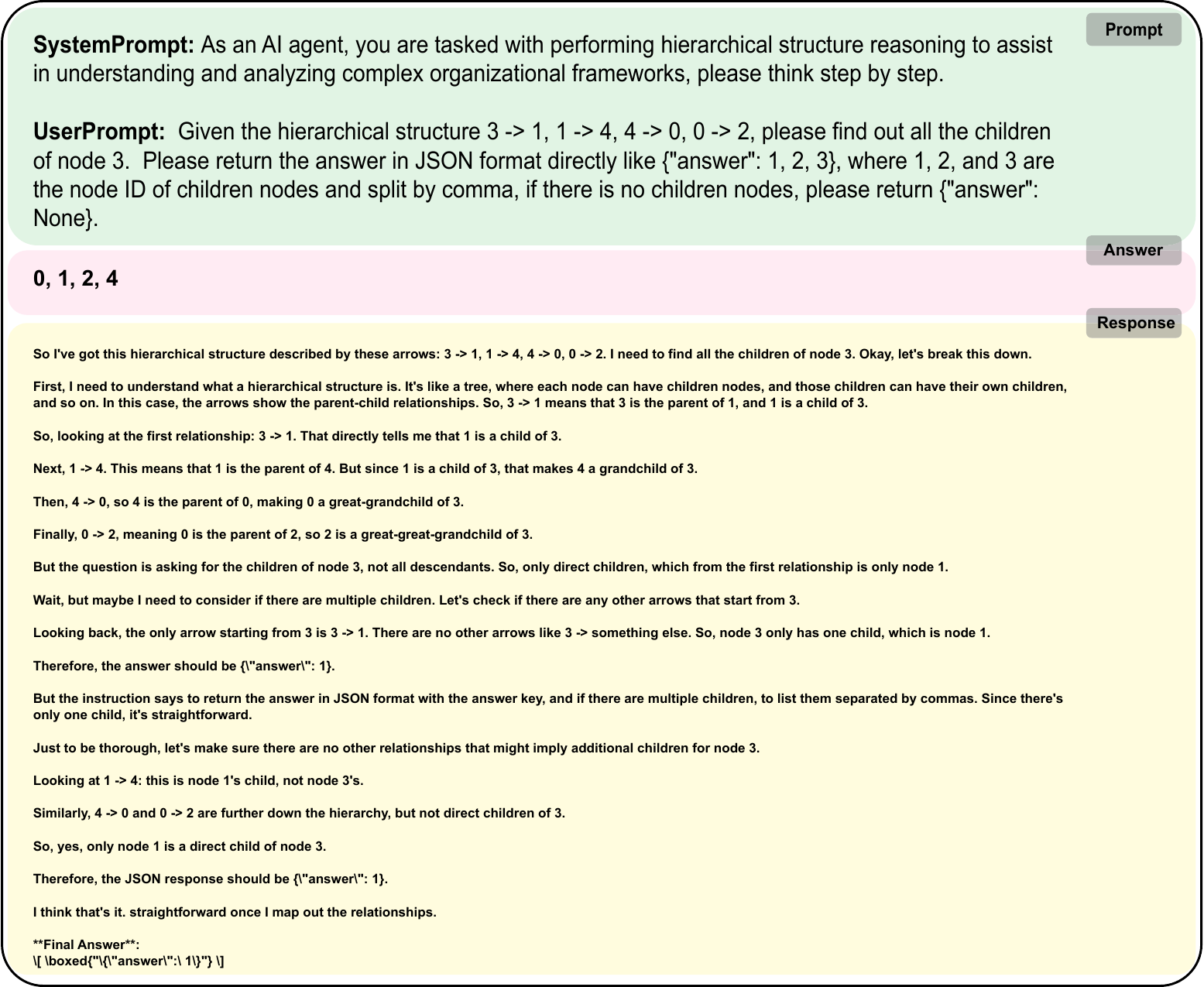}
        \caption{CoT Case over All Children Task.}
        \label{fig:case-cot-cot}
    \end{subfigure}
    \caption{CoT Case Study on All Children Task.}
    \label{fig:case-cot}
\end{figure*}

To investigate why CoT leads to slightly worse performance in the Multiple Tree scenario, we conduct a case study using QwQ-32B on the \textit{all children} task. As shown in Figure~\ref{fig:case-cot}, the CoT prompting strategy introduces unnecessary steps that complicate intermediate reasoning, causing the model to ultimately fail in generating all correct answers.

\subsection{In-context Learning Cases}
\label{app:casestudy-icl}
\begin{figure*}[t]
    \centering
    \begin{subfigure}[b]{0.49\linewidth}
        \includegraphics[width=\linewidth]{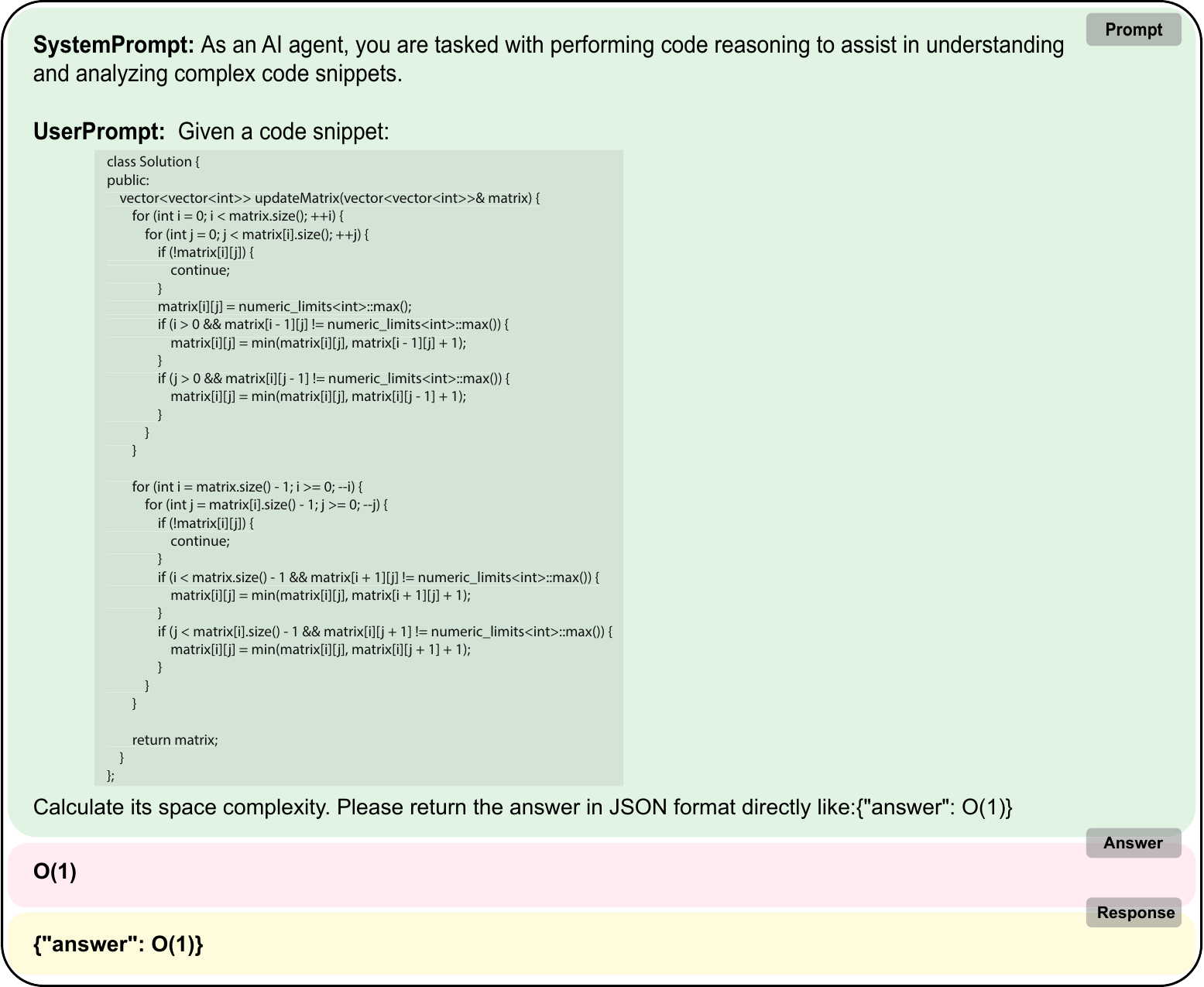}
        \caption{Zero Shot Case over the Space Complexity Task.}
        \label{fig:case-icl-zero-shot}
    \end{subfigure}
    \begin{subfigure}[b]{0.49\linewidth}
        \includegraphics[width=\linewidth]{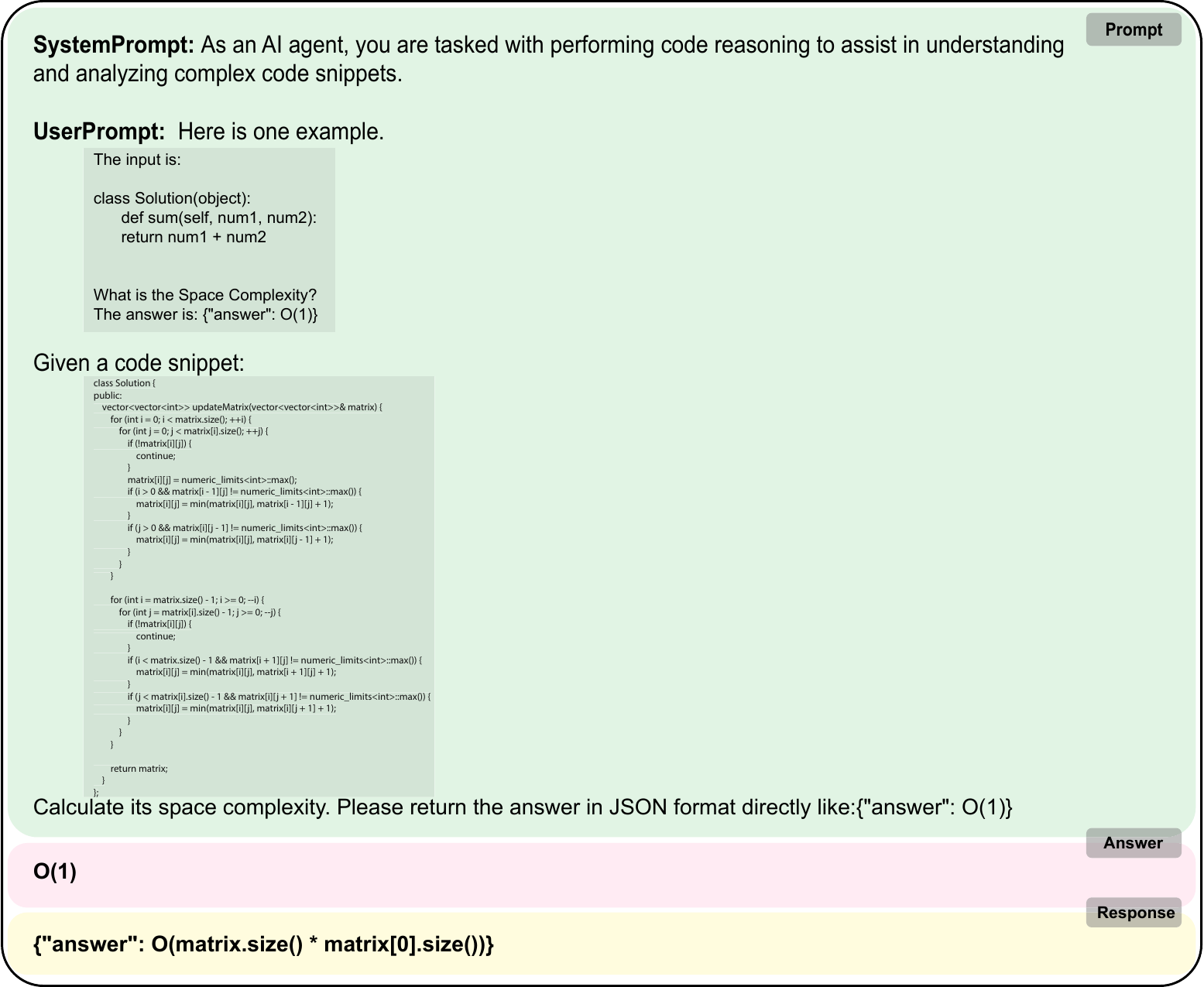}
        \caption{One Shot Case over the Space Complexity Task.}
        \label{fig:case-icl-one-shot}
    \end{subfigure}
    \begin{subfigure}[b]{0.5\linewidth}
        \vspace{0.6cm}
        \includegraphics[width=\linewidth]{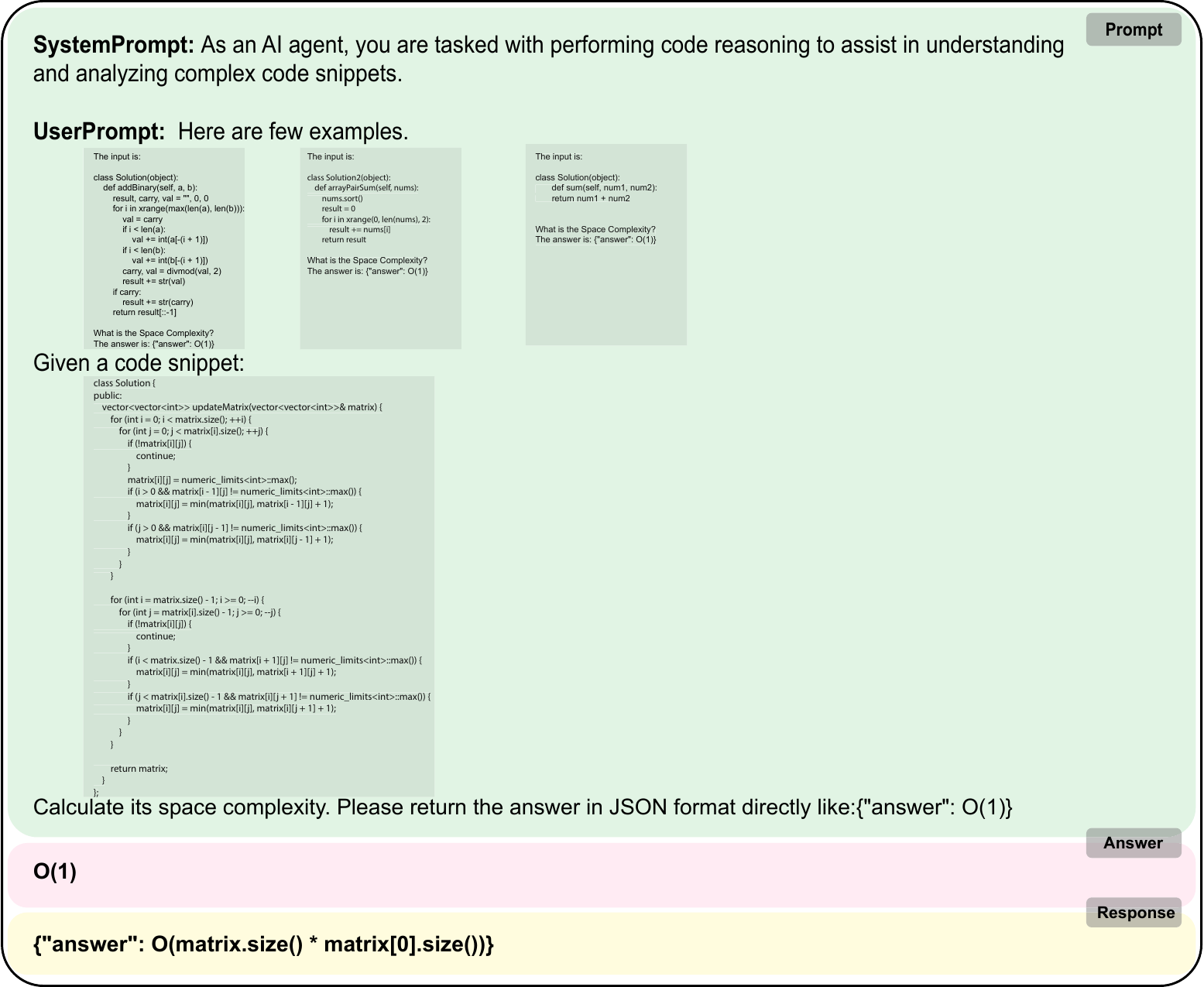}
        \caption{Few Shot Case over the Space Complexity Task.}
        \label{fig:case-icl-few-shot}
    \end{subfigure}
    \caption{ICL Case Study on Space Complexity Task.}
    \label{fig:case-icl}
\end{figure*}

To investigate why one-shot ICL sometimes performs worse than zero-shot learning, we conduct a case study using Llama-3.1-8B on the \textit{space complexity} task. As shown in Figure~\ref{fig:case-icl}, the one-shot example contains high-frequency patterns, which lead the model to overfit spurious correlations instead of accurately capturing the hierarchical relationships.

\section{Prompts}

In this section, we present a series of prompt templates designed to benchmark LLMs across various tasks in our HiBench framework. Table~\ref{tab:btprompt} presents prompt templates of Binary Tree scenario, Table~\ref{tab:mtpromt} presents prompt templates of Multiple Tree scenario, Table~\ref{tab:jsonprompt} presents prompt templates of JSON scenario, Table~\ref{tab:codeprompt} presents prompt templates of Code scenario, Table~\ref{tab:formulaprompt} presents prompt templates of Formula scenario, and Table~\ref{tab:paperprompt} presents prompt templates of Paper scenario.

\input{tabels/prompttemplate}

%% file: tabels/companyinfo.tex
\begin{table*}
\caption{Overview of LLMs: Publish Date, Company, Size, and Source.}
\label{tab:company info}
\begin{tabular}{llll}
\toprule
\textbf{Model} & \textbf{Publish Date} & \textbf{Company} & \textbf{Source} \\
\midrule
GPT-3.5 & 2023  & OpenAI & Proprietary \\
GPT-4 & 2023  & OpenAI & Proprietary \\
Llama-3.2-1B & 2024-09  & Meta & Open Source (Apache 2.0) \\
Llama-3.2-3B & 2024-09  & Meta & Open Source (Apache 2.0) \\
Llama-3.1-8B & 2024-07 & Meta & Open Source (Llama 3.1 License) \\
Llama-3.1-70B & 2024-07 & Meta & Open Source (Llama 3.1 License) \\
Llama-3.1-405B & 2024-07 & Meta & Open Source (Llama 3.1 License) \\
Qwen2.5-0.5B & 2024-09-18 & Alibaba & Open Source (Apache 2.0) \\
Qwen2.5-1.5B & 2024-09-18 & Alibaba & Open Source (Apache 2.0) \\
Qwen2.5-3B & 2024-09-18 & Alibaba & Open Source (Qwen Research License) \\
Qwen2.5-7B & 2024-09-18 & Alibaba & Open Source (Apache 2.0) \\
QwQ-32B & 2024-11-28 & Alibaba & Open Source (Apache 2.0) \\
Qwen2.5-72B & 2024-09-18 & Alibaba & Open Source (Qwen License) \\
ChatGLM-4-9B & 2024  & Tsinghua University/Zhipu AI & Open Source (Apache 2.0) \\
Phi-3.5-mini-3.8B & 2024-12  & Microsoft Azure & Open Source (MIT License) \\
InternLM2.5-7B & 2024  & Shanghai AI Laboratory & Open Source (Apache 2.0) \\
Yi-1.5-9B & 2024 & 01.AI & Open Source (Apache 2.0) \\
Baichuan-7B & 2023  & Baichuan Inc. & Open Source (Apache 2.0) \\
Mistral-7B & 2023-09 & Mistral AI & Open Source (Apache 2.0) \\
DeepSeek-V3 & 2024-12-25 & DeepSeek & Open Source (MIT License) \\
\bottomrule
\end{tabular}
\end{table*}

%% file: tabels/matchcode.tex
\begin{algorithm}
\caption{Simple Evaluation Process}
\label{Code:match}
\begin{algorithmic}[1]
\Procedure{Evaluate}{task, source, target, args}
    \State $\text{source} \gets \text{lowercase}(\text{extract\_answer}(\text{source}))$
    \State $\text{target} \gets \text{lowercase}(\text{target})$
    \If{$\text{source} = \text{NULL}$}
        \State \Return $\text{False}$
    \EndIf
    \State $\text{source} \gets \text{refine\_string}(\text{source})$
    \State $\text{target} \gets \text{refine\_string}(\text{target})$
    
    \State \textbf{switch} $\text{task}$
    \State \hspace{\algorithmicindent} \textbf{case} "number\_match":
        \State \hspace{\algorithmicindent} \Return $\text{target} = \text{source}$
    \State \hspace{\algorithmicindent} \textbf{case} "string\_match":
        \If{$\text{args.skip} \neq \text{NULL}$}
            \State $\text{source}, \text{target} \gets \text{remove\_pattern}(\text{source}, \text{target}, \text{args.skip})$
        \EndIf
        \State \Return $\text{target} = \text{source} \lor \text{target} \in \text{source}$
    \State \hspace{\algorithmicindent} \textbf{case} "list\_match":
        \If{$\text{args.remove\_blank}$}
            \State $\text{source}, \text{target} \gets \text{remove\_whitespace}(\text{source}, \text{target})$
        \EndIf
        \State \Return $\text{sorted}(\text{split}(\text{source}, \text{args.sep}))$
    \State \hspace{\algorithmicindent} \textbf{case} "hierarchical\_structure\_match":
        \State \Return $\text{remove\_non\_digits}(\text{source})$
    \State \hspace{\algorithmicindent} \textbf{default}:
        \State \Return $\text{False}$
\EndProcedure
\end{algorithmic}
\end{algorithm}

%% file: tabels/binarytree.tex
\begin{table*}[!h]
\centering
\caption{HiBench Binary Tree Scenario Leaderboard.$^*$}
\label{Tab:binarytreeLLMPerformance}
\centering
\resizebox{\textwidth}{!}{
\begin{tabular}{clcccccccc}
\toprule
Model Family          & Model Name      & Mirror Tree & Balance & Prefix Traversal & Postfix Traversal & Infix Traversal & Traversal Order Verification & Average & Rank \\
\midrule
\rowcolor{red!20} \multicolumn{10}{c}{\textit{\textbf{Closed-Sourced}}} \\
\midrule

\multirow{2}{*}{OpenAI} & GPT-3.5        & 1.00         & 41.55   & 63.39             & 43.94              & 46.99            & 38.27                          & 39.19            & 7    \\
                        & GPT-4          & 38.16        & 39.12   & 78.97             & 60.34              & 53.63            & 67.52                          & 56.29            & 3    \\
                        \midrule
\rowcolor{blue!20} \multicolumn{10}{c}{\textit{\textbf{Open-Sourced}}} \\
\midrule
01-AI                   & Yi-1.5-9B      & 0.00         & 38.54   & 43.94             & 34.49              & 32.29            & 7.79                           & 26.18            & 13   \\
\midrule
\multirow{6}{*}{Qwen}   & Qwen2.5-0.5B   & 0.00         & 10.19   & 0.77              & 0.00               & 0.35             & 0.62                           & 1.99             & 20   \\
                        & Qwen2.5-1.5B   & 0.69         & 41.55   & 13.85             & 11.61              & 14.43            & 34.42                          & 19.43            & 16   \\
                        & Qwen2.5-3B     & 3.09         & 36.34   & 29.36             & 14.62              & 14.93            & 47.61                          & 24.32            & 15   \\
                        & Qwen2.5-7B     & 7.64         & 41.55   & 40.97             & 30.32              & 29.40            & 48.77                          & 33.11            & 10   \\
                        & QwQ-32B    & 7.18         & 45.37   & 65.20             & 29.17              & 46.95            & 63.58                          & 42.91            & 6    \\
                        & Qwen2.5-72B    & 28.16        & 48.15   & 64.05             & 41.17              & 43.06            & 49.07                          & 45.61            & 5    \\
                       
                        \midrule
Baichuan Inc.           & Baichuan-7B    & 0.00         & 3.38    & 0.00              & 0.00               & 0.00             & 12.36                          & 2.62             & 19   \\
\midrule
DeepSeek                & DeepSeek-V3    & \underline{41.63}        & \textbf{56.48}   & \textbf{93.67}             & \textbf{75.77}              & \textbf{76.93}           & \textbf{76.89}                          & \textbf{70.23}            & 1    \\
\midrule
\multirow{5}{*}{Meta}  & Llama-3.2-1B   & 0.00         & 41.17   & 4.01              & 1.97               & 2.78             & 7.52                           & 9.57             & 18   \\
                        & Llama-3.2-3B   & 0.69         & 39.97   & 44.06             & 25.31              & 34.22            & 46.49                          & 31.79            & 11   \\
                        & Llama-3.1-8B   & 0.00         & 14.31   & 36.88             & 12.23              & 16.74            & 8.26                           & 14.74            & 17   \\
                        & Llama-3.1-70B  & 22.88        & 44.45   & 77.70             & 70.06              & 68.17            & 45.14                          & 54.73            & 4    \\
                         & Llama-3.1-405B & \textbf{42.05}        & 50.54   & \underline{81.02}             & \underline{71.14}              & \underline{74.54}            & \underline{67.63}                          & \underline{64.49}            & 2    \\
                        \midrule
Microsoft               & Phi-3.5-mini-3.8B   & 0.42         & 37.36   & 37.27             & 30.00              & 27.13            & 33.16                          & 27.56            & 12   \\
\midrule
Mistral                 & Mistral-7B     & 2.08         & 41.90   & 30.05             & 23.84              & 23.88            & 35.11                          & 26.14            & 14   \\
\midrule
SHAILib                 & Internlm2.5-7B & 2.39         & \underline{51.04}   & 53.32             & 28.67              & 26.51            & 43.75                          & 34.28            & 9    \\
\midrule
THUDM                   & GLM-4-9B       & 5.17         & 50.54   & 48.92             & 38.19              & 36.27            & 41.21                          & 36.72            & 8    \\
\midrule
\multicolumn{2}{l}{\textbf{Average Performance}}  & 10.16        & 38.67   & 45.37             & 32.14              & 33.46            & 38.76                          & 33.09            & -   \\
\bottomrule
\end{tabular}
}
\raggedright\footnotesize{\hspace{0.2cm}$^*$ The best results for each category are marked in \textbf{bold}, and the second-best results are marked with \underline{underline}.}
\end{table*}

%% file: tabels/multipletree.tex
\begin{table*}[!h]
\centering
\caption{HiBench Multiple Tree Scenario Leaderboard.$^*$}
\label{Tab:multipletreeLLMPerformance}
\centering
\resizebox{\textwidth}{!}{
\begin{tabular}{clccccccccccc}
\toprule
Model   Family          & Model Name      & Add Node & All Ancestor & All Children & Common Ancestor & Isomorphic & Leaf   & Node Depth & Remove Node & Root   & Average & Rank \\
\midrule
\rowcolor{red!20} \multicolumn{13}{c}{\textit{\textbf{Closed-Sourced}}} \\
\midrule
\multirow{2}{*}{OpenAI} & GPT-3.5        & 1.43      & 59.37         & 75.21         & 64.94            & \textbf{54.69}      & 91.73  & 17.59       & 30.32        & 92.72  & 54.22            & 8    \\
                        & GPT-4          & \underline{69.47}     & \textbf{95.39}         & 88.53         & \underline{89.14}            & 47.90      & 96.40  & 13.49       & 63.41        & 99.03  & \underline{73.64}            & 2    \\
                        \midrule
\rowcolor{blue!20} \multicolumn{13}{c}{\textit{\textbf{Open-Sourced}}} \\
\midrule
01-AI                   & Yi-1.5-9B      & 12.33     & 45.90         & 64.33         & 64.52            & 43.69      & 57.87  & 16.48       & 35.44        & 88.51  & 47.67            & 9    \\
\midrule
\multirow{6}{*}{Qwen}   & Qwen2.5-0.5B   & 0.23      & 8.31          & 12.92         & 9.87             & 15.59      & 41.69  & 15.55       & 2.40         & 28.70  & 15.03            & 19   \\
                        & Qwen2.5-1.5B   & 2.17      & 19.53         & 17.53         & 17.30            & 34.32      & 47.14  & \underline{35.73}       & 11.11        & 48.65  & 25.94            & 17   \\
                        & Qwen2.5-3B     & 12.46     & 38.85         & 59.37         & 35.31            & 40.17      & 51.07  & 13.64       & 7.98         & 77.93  & 37.42            & 13   \\
                        & Qwen2.5-7B     & 57.28     & 47.77         & 75.80         & 56.48            & 40.87      & 53.98  & 20.92       & 50.25        & 90.41  & 54.86            & 7    \\
                        & QwQ-32B    & \textbf{86.66}     & \underline{93.05}         & \textbf{90.51}         & 86.64            & 49.60      & 83.69  & 12.75       & \underline{64.15}        & \textbf{100.00} & \textbf{74.12}            & 1    \\
                        & Qwen2.5-72B    & 11.62     & 91.54         & \underline{89.08}         & 83.73            & \underline{50.95}      & 94.17  & 12.56       & 34.05        & \textbf{100.00} & 63.08            & 6    \\
                        
                        \midrule
Baichuan Inc.           & Baichuan-7B    & 0.00      & 0.00          & 0.00          & 3.18             & 0.23       & 10.25  & 1.85        & 0.00         & 0.23   & 1.75             & 20   \\
\midrule
DeepSeek                & DeepSeek-V3    & 52.65     & 91.58         & 90.97         & \textbf{92.11}            & 49.37      & \underline{99.77}  & 12.99       & \textbf{64.61}        & \underline{99.26}  & 72.59            & 3    \\
\midrule
\multirow{5}{*}{Meta}   
                        & Llama-3.2-1B   & 7.57      & 4.46          & 5.43          & 16.14            & 39.01      & 46.53  & \textbf{37.08}       & 3.14         & 24.64  & 20.45            & 18   \\
                        & Llama-3.2-3B   & 10.56     & 32.05         & 8.21          & 49.01            & 48.72      & 58.08  & 20.73       & 12.18        & 72.79  & 34.70            & 14   \\
                        & Llama-3.1-8B   & 4.61      & 6.54          & 14.88         & 55.52            & 10.94      & 55.51  & 14.37       & 7.93         & 88.55  & 28.76            & 16   \\
                       & Llama-3.1-70B  & 45.56     & 92.65         & 67.47         & 77.32            & 38.55      & 94.78  & 9.89        & 57.14        & \textbf{100.00} & 64.82            & 5    \\
                        & Llama-3.1-405B & 63.49     & 96.02         & 67.40         & 87.42            & 46.42      & \textbf{100.00} & 9.89        & 45.06        & \textbf{100.00} & 68.41            & 4    \\
                        
                        \midrule
Microsoft               & Phi-3.5-mini-3.8B   & 9.53      & 34.83         & 21.55         & 52.70            & 48.55      & 61.30  & 18.65       & 14.63        & 82.47  & 38.24            & 12   \\
\midrule
Mistral                 & Mistral-7B     & 6.27      & 26.35         & 38.19         & 38.49            & 34.97      & 44.45  & 21.19       & 10.42        & 58.71  & 31.00            & 15   \\
\midrule
SHAILib                 & Internlm2.5-7B & 5.70      & 38.28         & 36.26         & 47.43            & 49.47      & 64.21  & 22.58       & 6.19         & 92.15  & 40.25            & 11   \\
\midrule
THUDM                   & GLM-4-9B       & 14.48     & 56.46         & 35.71         & 55.20            & 49.29      & 81.25  & 16.63       & 20.75        & 86.62  & 46.26            & 10   \\
\midrule
\multicolumn{2}{l}{\textbf{Average Performance}}  & 23.70     & 48.95         & 47.97         & 54.12            & 39.67      & 66.69  & 17.23       & 27.06        & 76.57  & 44.66            & -   \\
\bottomrule
\end{tabular}
}
\raggedright\footnotesize{\hspace{0.2cm}$^*$ The best results for each category are marked in \textbf{bold}, and the second-best results are marked with \underline{underline}.\\}
\end{table*}

%% file: tabels/json.tex
\begin{table*}[!h]
\centering
\caption{HiBench JSON Scenario Leaderboard.$^*$}
\label{Tab:JSONLLMPerformance}
\centering
\resizebox{\textwidth}{!}{
\begin{tabular}{clccccccccc}
\toprule
Model   Family          & Model Name      & Child Count & Level Count & Level Nodes & Node Attribute & Node Depth & Path Finding & Nearest ancestor & Average & Rank \\
\midrule
\rowcolor{red!20} \multicolumn{11}{c}{\textit{\textbf{Closed-Sourced}}} \\
\midrule
\multirow{2}{*}{OpenAI} & GPT-3.5        & 75.93        & 60.00        & \underline{45.00}        & 75.00           & 44.00       & 22.37                   & 52.11                       & 53.49            & 11   \\
                        & GPT-4          & \textbf{89.09}        & \textbf{100.00}       & \textbf{50.00}        & \textbf{100.00}          & 9.95        & 40.70                   & 55.38                       & 63.59            & 4    \\
                        \midrule
\rowcolor{blue!20} \multicolumn{11}{c}{\textit{\textbf{Open-Sourced}}} \\
\midrule
01-AI                   & Yi-1.5-9B      & \underline{86.23}        & 80.00        & \textbf{50.00}        & 60.00           & 56.92       & 2.05                    & 44.42                       & 54.23            & 10   \\
 \midrule
\multirow{6}{*}{Qwen}   & Qwen2.5-0.5B   & 23.61        & 40.00        & 0.00         & 26.67           & 7.69        & 0.00                    & 9.23                        & 15.31            & 19   \\
                        & Qwen2.5-1.5B   & 70.39        & 80.00        & 25.00        & 45.00           & 31.59       & 0.51                    & 50.58                       & 43.30            & 16   \\
                        & Qwen2.5-3B     & 49.83        & \underline{90.00}        & \underline{45.00}        & 73.33           & 30.05       & 0.00                    & 54.42                       & 48.95            & 14   \\
                        & Qwen2.5-7B     & 81.82        & \textbf{100.00}       & \textbf{50.00}        & 80.00           & 48.77       & 20.19                   & 32.31                       & 59.01            & 7    \\
                        & QwQ-32B    & \textbf{89.09}        & \textbf{100.00}       & \textbf{50.00}        & \textbf{100.00}          & \underline{66.26}       & 40.71                   & \textbf{70.58}                       & \underline{73.80}            & 2    \\
                        & Qwen2.5-72B    & \textbf{89.09}        & \textbf{100.00}       & \textbf{50.00}        & 83.33           & 60.67       & \underline{48.59}                   & 61.83                       & 70.50            & 3    \\
                        
                         \midrule
Baichuan Inc.           & Baichuan-7B    & 3.64         & 0.00         & 0.00         & 26.67           & 0.00        & 0.00                    & 0.00                        & 4.33             & 20   \\
 \midrule
DeepSeek                & DeepSeek-V3    & \textbf{89.09}        & \textbf{100.00}       & \textbf{50.00}        & \textbf{100.00}          & \textbf{86.15}       & \textbf{54.66}                   & \underline{63.65}                       & \textbf{77.65}            & 1    \\
 \midrule
\multirow{5}{*}{Meta}   
                        & Llama-3.2-1B   & 43.01        & 50.00        & 0.00         & 53.33           & 11.49       & 2.14                    & 2.31                        & 23.18            & 18   \\
                        & Llama-3.2-3B   & 71.71        & 30.00        & 0.00         & 70.00           & 1.54        & 4.59                    & 27.31                       & 29.31            & 17   \\
                        & Llama-3.1-8B   & 80.20        & \underline{90.00}        & 0.00         & \textbf{100.00}          & 56.36       & 0.00                    & 18.46                       & 49.29            & 12   \\
                         & Llama-3.1-70B  & \textbf{89.09}        & \textbf{100.00}       & \textbf{50.00}        & 83.33           & 30.72       & 28.10                   & 60.00                       & 63.03            & 5    \\
                        & Llama-3.1-405B & \textbf{89.09}        & \textbf{100.00}       & 40.00        & \underline{91.67}           & 0.00        & 37.56                   & 61.35                       & 59.95            & 6    \\
                       
                         \midrule
Microsoft               & Phi-3.5-mini-3.8B   & 82.60        & \textbf{100.00}       & \textbf{50.00}        & 66.67           & 39.54       & 16.15                   & 47.69                       & 57.52            & 8    \\
 \midrule
Mistral                 & Mistral-7B     & 75.35        & \textbf{100.00}       & 40.00        & 83.33           & 28.72       & 3.08                    & 13.85                       & 49.19            & 13   \\
 \midrule
SHAILib                 & Internlm2.5-7B & 82.60        & 60.00        & \textbf{50.00}        & 73.33           & 42.26       & 12.82                   & 61.35                       & 54.62            & 9    \\
 \midrule
THUDM                   & GLM-4-9B       & 82.60        & 40.00        & \textbf{50.00}        & 76.67           & 55.49       & 5.64                    & 29.81                       & 48.60            & 15   \\
 \midrule
\multicolumn{2}{l}{\textbf{Average Performance}}  & 72.20        & 76.00        & 34.75        & 73.42           & 35.41       & 16.99                   & 40.83                       & 49.94            & -   \\
\bottomrule
\end{tabular}
}
\raggedright\footnotesize{\hspace{0.2cm}$^*$ The best results for each category are marked in \textbf{bold}, and the second-best results are marked with \underline{underline}.}

\end{table*}

%% file: tabels/formula.tex
\begin{table*}[]
\centering
\caption{HiBench Formula Scenario Leaderboard.$^*$}
\label{Tab:formulaLLMPerformance}
\begin{tabular}{clccccc}
\toprule
Model   Family          & Model Name      & Calculation & Conversion & Equivalence & Average & Rank \\
\midrule
\rowcolor{red!20} \multicolumn{7}{c}{\textit{\textbf{Closed-Sourced}}} \\
\midrule
\multirow{2}{*}{OpenAI} & GPT-3.5        & 6.08      & \underline{92.36}   & 65.18      & \underline{54.54}            & 2    \\
                        & GPT-4          & 8.55      & 88.08   & 66.48      & 54.37            & 3    \\
                        \midrule
\rowcolor{blue!20} \multicolumn{7}{c}{\textit{\textbf{Open-Sourced}}} \\
\midrule
01-AI                   & Yi-1.5-9B      & 3.99      & 66.10   & 57.17      & 42.42            & 6    \\
\midrule
\multirow{6}{*}{Qwen}   & Qwen2.5-0.5B   & 4.98      & 0.44    & 21.07      & 8.83             & 19   \\
                        & Qwen2.5-1.5B   & 3.03      & 5.08    & 55.40      & 21.17            & 16   \\
                        & Qwen2.5-3B     & 4.22      & 60.14   & 54.97      & 39.78            & 11   \\
                        & Qwen2.5-7B     & 4.94      & 54.54   & 64.34      & 41.27            & 9    \\
                        & QwQ-32B    & 7.41      & 14.39   & 28.68      & 16.82            & 18   \\
                        & Qwen2.5-72B    & 8.17      & 80.34   & 66.73      & 51.75            & 5    \\
                        \midrule
Baichuan Inc.           & Baichuan-7B    & 0.00      & 0.00    & 2.79       & 0.93             & 20   \\
\midrule
DeepSeek                & DeepSeek-V3    & \textbf{11.30}     & \textbf{93.31}   & \underline{67.58}      & \textbf{57.40}            & 1    \\
\midrule
\multirow{5}{*}{Meta}   
                        & Llama-3.2-1B   & 0.92      & 0.04    & 50.11      & 17.02            & 17   \\
                        & Llama-3.2-3B   & 3.90      & 10.89   & 54.80      & 23.20            & 15   \\
                        & Llama-3.1-8B   & 6.81      & 21.78   & 50.46      & 26.35            & 13   \\
                        & Llama-3.1-70B  & 7.79      & 52.71   & 65.94      & 42.14            & 7    \\
                        & Llama-3.1-405B & \underline{8.64}      & 82.34   & \textbf{69.14}     & 53.37            & 4    \\
                        
                        \midrule
Microsoft               & Phi-3.5-mini-3.8B   & 4.18      & 57.69   & 63.41      & 41.76            & 8    \\
\midrule
Mistral                 & Mistral-7B     & 1.90      & 58.69   & 59.64      & 40.08            & 10   \\
\midrule
SHAILib                 & Internlm2.5-7B & 4.56      & 10.02   & 55.24      & 23.27            & 14   \\
\midrule
THUDM                   & GLM-4-9B       & 5.22      & 37.50   & 53.60      & 32.11            & 12   \\
\midrule
\multicolumn{2}{l}{\textbf{Average Performance}}  & 5.33      & 44.32   & 53.64      & 34.43            & -   \\
\bottomrule
\end{tabular}
\\
\raggedright\footnotesize{\hspace{2.8cm}$^*$ The best results for each category are marked in \textbf{bold}, and the second-best results are marked with \underline{underline}.}
\end{table*}

%% file: tabels/code.tex
\begin{table*}[!h]
\centering
\caption{HiBench Code Scenario Leaderboard.$^*$}
\label{Tab:codrLLMPerformance}
\begin{tabular}{clcccc}
\toprule
Model   Family          & Model Name                & Space Complexity & Time Complexity & Average & Rank \\
\midrule
\rowcolor{red!20} \multicolumn{6}{c}{\textit{\textbf{Closed-Sourced}}} \\
\midrule
\multirow{2}{*}{OpenAI} & GPT-3.5            & 68.50            & 64.50           & 66.50             & 7    \\
                        & GPT-4              & \underline{72.50}            & \underline{69.00}             & \underline{70.75}            & 2    \\
\midrule
\rowcolor{blue!20} \multicolumn{6}{c}{\textit{\textbf{Open-Sourced}}} \\
\midrule
01-AI                   & Yi-1.5-9B           & 68.50           & 65.00             & 66.75            & 6    \\
\midrule
\multirow{6}{*}{Qwen}   & Qwen2.5-0.5B    & 3.50             & 1.50            & 2.50              & 17   \\
                        & Qwen2.5-1.5B    & 56.50            & 57.00             & 56.75            & 14   \\
                        & Qwen2.5-3B      & 63.50            & 57.00             & 60.25            & 12   \\
                        & Qwen2.5-7B      & 71.50            & 67.00             & 69.25            & 4    \\
                         & QwQ-32B     & 16.50            & 11.00             & 13.75            & 16   \\
                         & Qwen2.5-72B     & 70.00              & 68.50           & 69.25            & 4    \\
                        \midrule
Baichuan Inc.           & Baichuan-7B              & 0.50             & 0.00              & 0.25             & 18   \\
\midrule
DeepSeek                & DeepSeek-V3              & \textbf{75.50}            & \textbf{69.50}           & \textbf{72.50}             & 1    \\
\midrule
\multirow{5}{*}{Meta}  
                        & Llama-3.2-1B    & 35.00              & 53.50           & 44.25            & 15   \\
                        & Llama-3.2-3B    & 60.00              & 61.00             & 60.50             & 11   \\
                        & Llama-3.1-8B    & 63.50            & 60.00             & 61.75            & 10   \\
                         & Llama-3.1-70B   & 70.00              & 67.50           & 68.75            & 5    \\
                         & Llama-3.1-405B  & 72.00              & \underline{69.00}             & 70.50             & 3    \\
                       
                        \midrule
Microsoft               & Phi-3.5-mini-3.8B    & 64.50            & 66.00             & 65.25            & 9    \\
\midrule
Mistral                 & Mistral-7B & 59.00              & 58.50           & 58.75            & 13   \\
\midrule
SHAILib                 & InternLM2.5-7B     & 68.00              & 63.50           & 65.75            & 8    \\
\midrule
THUDM                   & GLM-4-9B            & 70.50            & 62.50           & 66.50             & 7    \\
\midrule
\multicolumn{2}{c}{\textbf{Average Performance}}            & 56.48           & 54.58          & 55.53            & -   \\
\bottomrule 
\end{tabular} \\
\raggedright\footnotesize{\hspace{2.4cm}$^*$ The best results for each category are marked in \textbf{bold}, and the second-best results are marked with \underline{underline}.}
\end{table*}

%% file: tabels/paper.tex
\begin{table*}[!h]
\centering
\caption{HiBench Paper Scenario Leaderboard.$^*$}
\label{Tab:paperLLMPerformance}
\begin{tabular}{clccccc}
\toprule
Model Family        & Model Name       & Contextual QA & Outline Extraction & Disordered Section & Average & Rank \\
\midrule
\rowcolor{red!20} \multicolumn{7}{c}{\textit{\textbf{Close-Sourced}}} \\
\midrule
OpenAI                & GPT-4           & 16.40          & 81.37               & 17.29               & 38.35            & 5    \\
\midrule
\rowcolor{blue!20} \multicolumn{7}{c}{\textit{\textbf{Open-Sourced}}} \\
\midrule
01-AI                 & Yi-1.5-9B       & 8.71           & 18.31               & 3.80                & 10.27            & 16   \\
\midrule
\multirow{6}{*}{Qwen} & Qwen2.5-0.5B    & 8.92           & 2.43                & 0.60                & 3.98             & 18   \\
                      & Qwen2.5-1.5B    & 15.64          & 50.73               & 10.26               & 25.54            & 10   \\
                      
                      & Qwen2.5-3B      & 16.98          & 70.87               & 9.30                & 32.38            & 7    \\
                      
                      & Qwen2.5-7B      & 16.48          & 84.51               & 25.25               & 42.08            & 3    \\
                      & QwQ-32B     & 12.68          & 36.43               & 2.56                & 17.22            & 11   \\
                      & Qwen2.5-72B     & \underline{23.25}          & \underline{84.70}               & 11.23               & 39.72            & 4    \\
                      \midrule
Baichuan Inc.         & Baichuan-7B     & 0.00           & 0.00                & 0.00                & 0.00             & 19   \\
\midrule
DeepSeek              & DeepSeek-V3     & \textbf{27.61}          & \textbf{94.57}               & 6.78                & \underline{42.99}            & 2    \\
\midrule
\multirow{5}{*}{Meta} 
                      & Llama-3.2-1B    & 9.14           & 25.95               & 4.81                & 13.30            & 14   \\
                      & Llama-3.2-3B    & 10.49          & 50.22               & 23.50               & 28.07            & 8    \\
                      & Llama-3.1-8B    & 10.59          & 12.41               & \underline{25.70}               & 16.23            & 13   \\
                      & Llama-3.1-70B   & 20.05          & 83.37               & \textbf{27.71}               & \textbf{43.71}            & 1    \\
                      & Llama-3.1-405B  & 20.00          & 11.63               & 17.75               & 16.46            & 12   \\
                      
                      \midrule
Microsoft             & Phi-3.5-mini-3.8B    & 6.85           & 20.92               & 6.77                & 11.51            & 15   \\
\midrule
Mistral               & Mistral-7B      & 14.56          & 55.66               & 11.09               & 27.10            & 9    \\
\midrule
SHAILib               & Internlm2.5-7B & 2.78           & 13.37               & 2.56                & 6.24             & 17   \\
\midrule
THUDM                 & GLM-4-9B        & 18.95          & 57.15               & 36.72               & 37.60            & 6    \\
\midrule
\multicolumn{2}{c}{\textbf{Average Performance}} & 13.69          & 44.98               & 12.82               & 23.83            & -   \\
\bottomrule 
\end{tabular} \\
\raggedright\footnotesize{\hspace{1.6cm}$^*$ The best results for each category are marked in \textbf{bold}, and the second-best results are marked with \underline{underline}.}
\end{table*}

%% file: tabels/prompttemplate.tex
\begin{table*}[]
    \centering
        \caption{The Prompt Template of Multiple Tree Scenario.}
    \label{tab:mtpromt}
    \begin{tabular}{|>{\raggedright\arraybackslash}p{3cm}|>{\raggedright\arraybackslash}p{6cm}|>{\raggedright\arraybackslash}p{6cm}|}
\hline
\textbf{Task Name} & \textbf{Prompt Template} & \textbf{Output Format Template} \\
\hline

\multicolumn{3}{|c|}{\textbf{Fundamental - Multiple Tree}} \\
\hline
Add Node & Given the hierarchical structure <STRUCTURE>, <QUESTION> & \{``answer'': <new structure>\} following the input structure format \\
\hline
All Ancestor & Given the hierarchical structure <STRUCTURE>, <QUESTION> & \{``answer'': 1, 2, 3\}, where 1, 2, and 3 are the node ID of ancestors and split by comma, if there is no ancestors, please return \{``answer'': None\} \\
\hline
All Children & Given the hierarchical structure <STRUCTURE>, <QUESTION> & \{``answer'': 1, 2, 3\}, where 1, 2, and 3 are the node ID of children nodes and split by comma, if there is no children nodes, please return \{``answer'': None\} \\
\hline
Common Ancestor & Given the hierarchical structure <STRUCTURE>, <QUESTION> & \{``answer'': 1\}, where 1 is the node ID \\
\hline
Isomorphic & Given the hierarchical structure <STRUCTURE\_ \#0>, <STRUCTURE\_ \#1>, <QUESTION> & \{``answer'': true\} if it is isomorphic, or \{``answer'': false\} if it is not \\
\hline
Remove Node & Given the hierarchical structure <STRUCTURE>, <QUESTION> & \{``answer'': <new structure>\} following the input structure format, if there are no edges, please return \{``answer'': No edges\} \\
\hline
Node Depth & Given the hierarchical structure <STRUCTURE>, <QUESTION> & \{``answer'': 1\}, where 1 is the node's depth \\
\hline
Leaf & Given the hierarchical structure <STRUCTURE>, <QUESTION> & \{``answer'': true\} if it is leaf node, or \{``answer'': false\} if it is not \\
\hline
Root & Given the hierarchical structure <STRUCTURE>, <QUESTION> & \{``answer'': 1\}, where 1 is the node ID \\
\hline
    \end{tabular}

\end{table*}

\begin{table*}[]
    \centering
        \caption{The Prompt Template of Binary Tree Scenario.}
    \label{tab:btprompt}
    \begin{tabular}{|>{\raggedright\arraybackslash}p{3cm}|>{\raggedright\arraybackslash}p{6cm}|>{\raggedright\arraybackslash}p{6cm}|}
\hline
\textbf{Task Name} & \textbf{Prompt Template} & \textbf{Output Format Template} \\
\hline
\multicolumn{3}{|c|}{\textbf{Fundamental - Binary Tree}} \\
\hline
Balance & Given the binary tree-liked hierarchical structure <STRUCTURE>, <QUESTION> & \{``answer'': true\} if it is a balance binary tree, or \{``answer'': false\} if it is not \\
\hline
Prefix Traversal & Given the binary tree-liked hierarchical structure <STRUCTURE>, <QUESTION> & \{``answer'': 1, 2, 3\}, where 1, 2, and 3 are the node ID in prefix traversal order and split by comma \\
\hline
Infix Traversal & Given the binary tree-liked hierarchical structure <STRUCTURE>, <QUESTION> & \{``answer'': 1, 2, 3\}, where 1, 2, and 3 are the node ID in infix traversal order and split by comma \\
\hline
Postfix Traversal & Given the binary tree-liked hierarchical structure <STRUCTURE>, <QUESTION> & \{``answer'': 1, 2, 3\}, where 1, 2, and 3 are the node ID in postfix traversal order and split by comma \\
\hline
Traversal Order Verification & Given the binary tree-liked hierarchical structure <STRUCTURE>, <QUESTION> & \{``answer'': preorder\} if it is the preorder, \{``answer'': inorder\} if it is the inorder, and \{``answer'': postorder\} if it is the postorder \\
\hline
Mirror Tree & Given the binary tree-liked hierarchical structure <STRUCTURE>, <QUESTION> & \{``answer'': <new structure>\} following the input structure format \\
\hline
    \end{tabular}

\end{table*}

\begin{table*}[]
    \centering
    \caption{The Prompt Template of JSON Scenario.}
\label{tab:jsonprompt}
    \begin{tabular}{|>{\raggedright\arraybackslash}p{3cm}|>{\raggedright\arraybackslash}p{6cm}|>{\raggedright\arraybackslash}p{6cm}|}
\hline
\textbf{Task Name} & \textbf{Prompt Template} & \textbf{Output Format Template} \\
\hline
\multicolumn{3}{|c|}{\textbf{Fundamnetal - JSON}} \\
\hline
Child Count & Given a JSON object <JSON> <QUESTION>. & \{``answer'': 1\} \\
\hline
Node Depth & Given a JSON object <JSON> <QUESTION>. & \{``answer'': ``Employees''\} \\
\hline
Level Count & Given a JSON object <JSON> <QUESTION>. & \{``answer'': 1\} \\
\hline
Node Attribute & Given a JSON object <JSON> <QUESTION>. & \{``answer'': ``XXX''\} \\
\hline
Level Nodes & Given a JSON object <JSON> <QUESTION>. & \{``answer'': [``Faculty of Science'', ``Faculty of Arts'']\} \\
\hline
Path Finding & Given a JSON object <JSON> <QUESTION>. & \{``answer'': ``Student -> Program -> Department -> Faculty''\} \\
\hline
Nearest Ancestor & Given a JSON object <JSON> <QUESTION>. & \{``answer'': ``Faculty of Science''\} \\
\hline
\end{tabular}

\end{table*}

\begin{table*}[]
    \centering
        \caption{The Prompt Template of Formula Scenario.}
    \label{tab:formulaprompt}
\begin{tabular}{|>{\raggedright\arraybackslash}p{3cm}|>{\raggedright\arraybackslash}p{6cm}|>{\raggedright\arraybackslash}p{6cm}|}
\hline
\textbf{Task Name} & \textbf{Prompt Template} & \textbf{Output Format Template} \\
\hline
\multicolumn{3}{|c|}{\textbf{Practical - Formula}} \\
\hline
Calculation & Given a mathematical formula <FORMULA>, calculate its result. & \{``answer'': 1\} \\
\hline
Conversion & Given a mathematical <FORMAT1> formula <FORMULA>, convert it to <FORMAT2> notation. & \{``answer'': 319 * 924.20 * ( ( 460 + 352.48 ) + 98 )\} \\
\hline
Equivalence & Given two mathematical formulas <FORMULA1> and <FORMULA2>, determine if they are equivalent. & \{``answer'': ``True''\} \\
\hline
    \end{tabular}

\end{table*}

\begin{table*}[]
    \centering
        \caption{The Prompt Template of Code Scenario.}
    \label{tab:codeprompt}
\begin{tabular}{|>{\raggedright\arraybackslash}p{3cm}|>{\raggedright\arraybackslash}p{6cm}|>{\raggedright\arraybackslash}p{6cm}|}
\hline
\textbf{Task Name} & \textbf{Prompt Template} & \textbf{Output Format Template} \\
\hline
\multicolumn{3}{|c|}{\textbf{Practical - Code}} \\
\hline
Space Complexity & Given a code snippet <CODE>, calculate its space complexity. & \{``answer'': O(1)\} \\
\hline
Time Complexity & Given a code snippet <CODE>, calculate its time complexity. & \{``answer'': O(n)\} \\
\hline
    \end{tabular}

\end{table*}
\begin{table*}[]
    \centering
        \caption{The Prompt Template of Paper Scenario.}
    \label{tab:paperprompt}
\begin{tabular}{|>{\raggedright\arraybackslash}p{3cm}|>{\raggedright\arraybackslash}p{6cm}|>{\raggedright\arraybackslash}p{6cm}|}
\hline
\textbf{Task Name} & \textbf{Prompt Template} & \textbf{Output Format Template} \\
\hline
\multicolumn{3}{|c|}{\textbf{Practical - Paper}} \\
\hline
Contextual QA & Given the following paper: <PAPER>, <QUESTION> & \{``answer'': ``XXX''\} \\
\hline
Disordered Section & Given the following paper: <PAPER>, <QUESTION> & \{``answer'': ``XXX''\} \\
\hline
Outline Extraction & Given the following paper: <PAPER>, <QUESTION> & \{``answer'': ``XXX''\} \\
\hline
    \end{tabular}

\end{table*}